\documentclass[runningheads]{llncs}

\usepackage{eccv}

\usepackage{eccvabbrv}

\usepackage{graphicx}
\usepackage{booktabs}
\usepackage{tabularx}
\usepackage{amsmath}
\usepackage{stmaryrd}
\usepackage{multirow}
\usepackage{array}

\usepackage[accsupp]{axessibility}  

\usepackage{hyperref}

\usepackage{orcidlink}

\begin{document}

\title{360U-Former: HDR Illumination Estimation with Panoramic Adapted Vision Transformers} 
\titlerunning{360U-Former}

\author{Jack Hilliard\inst{1,2}\orcidlink{0009-0008-0906-1295} \and
Adrian Hilton\inst{1,3}\orcidlink{0000-0003-4223-238X} \and
Jean-Yves Guillemaut\inst{1,4}\orcidlink{0000-0001-8223-5505}}

\authorrunning{J.~Hilliard et al.}

\institute{Unviersity of Surrey, Guildford, Surrey, UK \and
\email{jh00695@surrey.ac.uk} \and
\email{a.hilton@surrey.ac.uk} \and
\email{j.guillemaut@surrey.ac.uk}}

\maketitle

\begin{abstract}
  Recent illumination estimation methods have focused on enhancing the resolution and improving the quality and diversity of the generated textures. However, few have explored tailoring the neural network architecture to the Equirectangular Panorama (ERP) format utilised in image-based lighting. Consequently, high dynamic range images (HDRI) results usually exhibit a seam at the side borders and textures or objects that are warped at the poles. To address this shortcoming we propose a novel architecture, 360U-Former, based on a U-Net style Vision-Transformer which leverages the work of PanoSWIN, an adapted shifted window attention tailored to the ERP format. To the best of our knowledge, this is the first purely Vision-Transformer model used in the field of illumination estimation. We train 360U-Former as a GAN to generate HDRI from a limited field of view low dynamic range image (LDRI). We evaluate our method using current illumination estimation evaluation protocols and datasets, demonstrating that our approach outperforms existing and state-of-the-art methods without the artefacts typically associated with the use of the ERP format.
\end{abstract}

\section{Introduction}
\label{sec:intro}

To believably composite an object into a virtual scene it must be lit consistently with the target scene's illumination conditions. The field of illumination estimation has sought to capture a scene’s lighting through the constraints of a Limited Field of View (LFOV) Low Dynamic Range image (LDR(I)) such as that taken from a mobile phone camera.
Several methods have been utilised to represent these conditions such as regression-based methods, like Spherical Harmonics (SH) \cite{Green2003} and Spherical Gaussians (SG) \cite{Tsai2006}, and Image Based Lighting (IBL) \cite{Debevec1998} methods that render lighting from an Equirectangular Panorama (ERP) High Dynamic Range image (HDR(I)). IBL has become the leading way to represent lighting conditions \cite{Legendre2019, Somanath2021, Weber2022, Wang2022, Dastjerdi2023, Hilliard2023} due to its ability to capture high-frequency textures, as well as global illumination, meaning it can be used to light a range of surfaces from rough diffuse to mirror reflective. The current trends in IBL and illumination estimation are to increase the resolution, details and accuracy of the generated ERP images \cite{Weber2022, Wang2022, Dastjerdi2023, Hilliard2023}.

\begin{figure*}[t]
    \centering
    \setlength\tabcolsep{0pt}
    \renewcommand{\arraystretch}{0.5}
        \begin{subfigure}[b]{0.24\linewidth}
		\centering
		\centerline{\includegraphics[width=\textwidth]{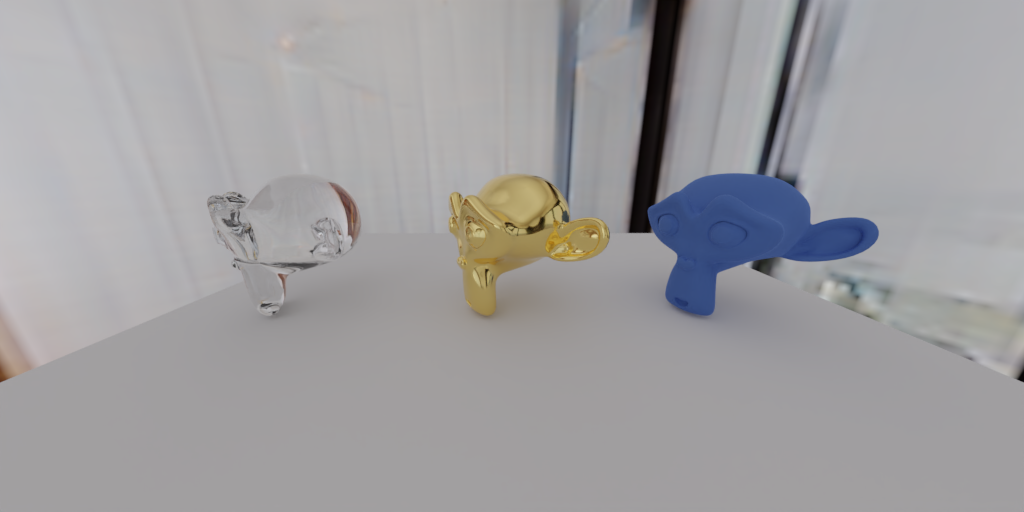}}
        \vspace{0.3ex}
        \centerline{\includegraphics[width=\textwidth]{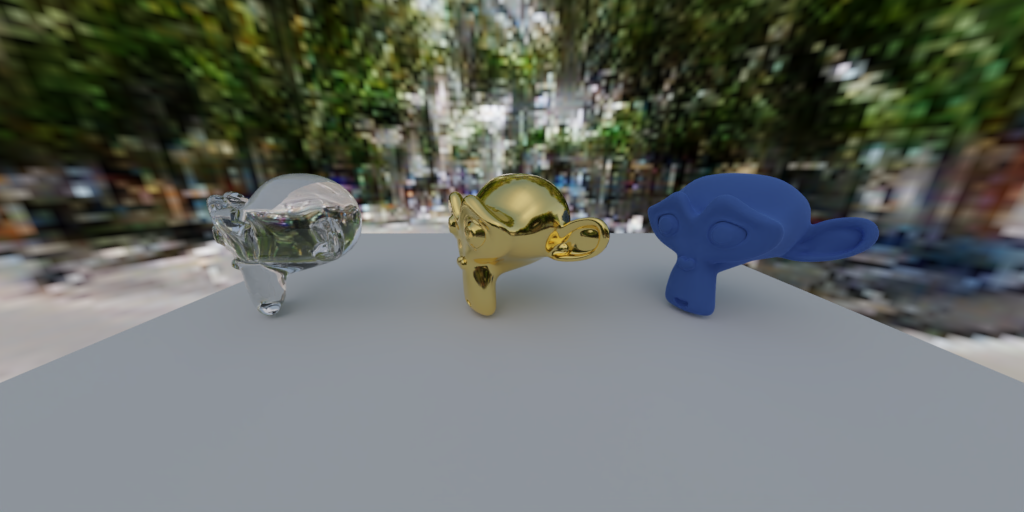}}
		\caption{\scriptsize{360U-Former render}}\medskip
        \label{fig:render-a}
	\end{subfigure}
	\begin{subfigure}[b]{0.24\linewidth}
		\centering
        \centerline{\includegraphics[width=\textwidth]{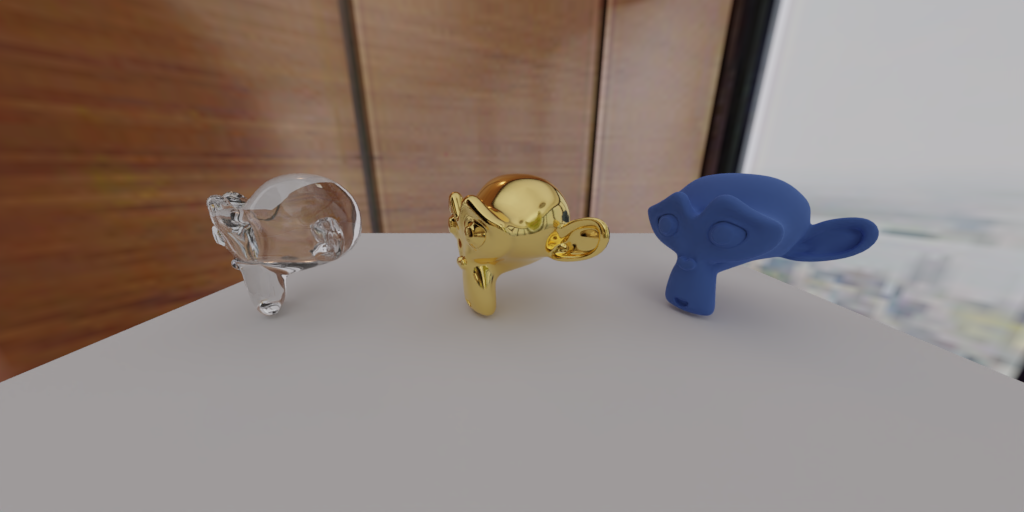}}
        \vspace{0.3ex}
        \centerline{\includegraphics[width=\textwidth]{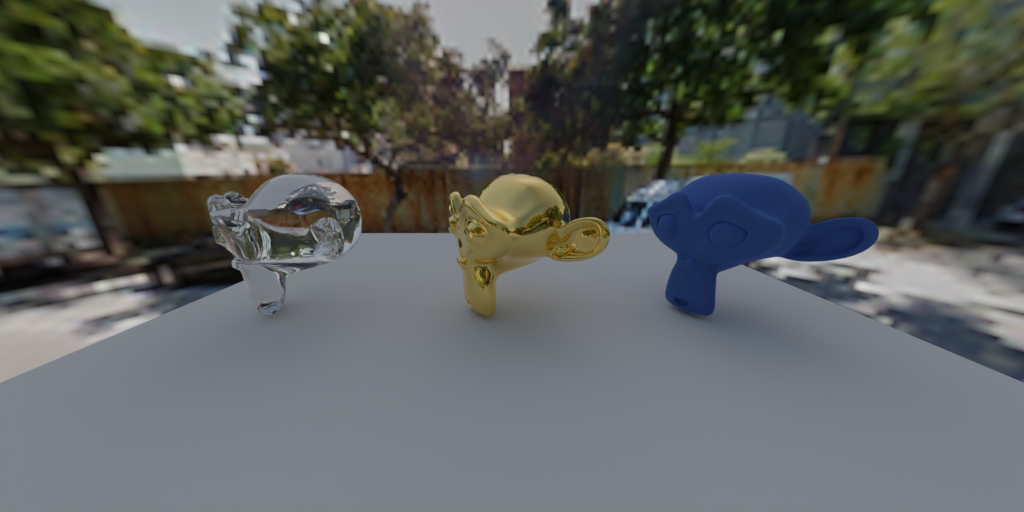}}
		\caption{\scriptsize{GT render}}\medskip
        \label{fig:render-b}
	\end{subfigure}
	\begin{subfigure}[b]{0.24\linewidth}
		\centering
		\centerline{\includegraphics[width=\textwidth]{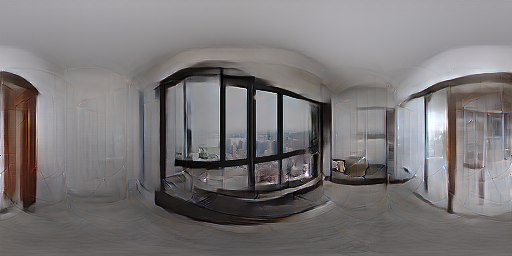}}
        \vspace{0.3ex}
        \centerline{\includegraphics[width=\textwidth]{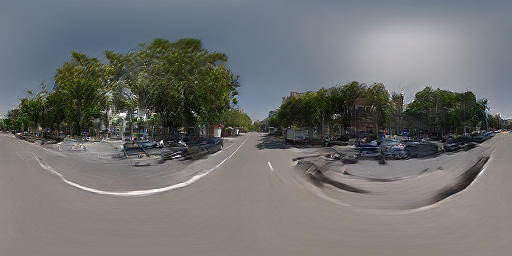}}
		\caption{\scriptsize{360U-Former EM}}\medskip
        \label{fig:render-c}
	\end{subfigure}
        \begin{subfigure}[b]{0.24\linewidth}
		\centering
		\centerline{\includegraphics[width=\textwidth]{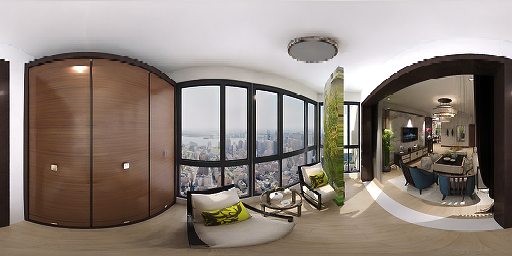}}
        \vspace{0.3ex}
        \centerline{\includegraphics[width=\textwidth]{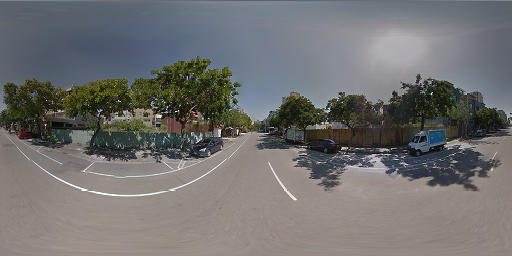}}
		\caption{\scriptsize{GT EM}}\medskip
        \label{fig:render-d}
	\end{subfigure}
\caption{Examples of an object with different surface properties being rendered with an HDRI environment map (EM) of an indoor ({\it Top}) and outdoor ({\it Bottom}) scene, from either the ground truth or generated by our network 360U-Former with PanoSWIN attention blocks. We also include the EM for each scene and method for reference.}
\label{fig:object_render}
\end{figure*}

Neural networks have used the ERP image format in various tasks, such as panoramic outpainting and illumination estimation. Due to mapping the surface of a sphere to a 2D image, warping occurs at the top and bottom of ERP images. When using ERP images in a standard neural network this warping has to be learnt by the network or accounted for by adapting the architecture. The sides of the image also have to be considered connected, unlike regular LFOV images. Not adapting a neural network architecture to work with the ERP format can lead to artefacts such as obvious seams at the side borders of the image and badly generated objects at the top and bottom (poles) of the ERP where it is most warped. In the field of illumination estimation, only a few models have made adjustments \cite{Somanath2021, Dastjerdi2023, Akimoto2022, Dastjerdi2022}. These few approaches are typically limited to either changes to the loss functions or changes to the image at network inference.

Vision Transformers (ViT) have shown their ability to understand relationships of both global and local information in an image for a variety of image processing tasks such as object classification \cite{Ling2023}, image restoration \cite{Deng2022}  and outpainting \cite{Gao2023}, as well as being able to change the window attention network to better process ERPs \cite{Ling2023}.

We propose 360U-Former, a U-Net style ViT adapted for the ERP format by leveraging PanoSWIN \cite{Ling2023} attention blocks.
We use 360U-Former to generate the ERP HDR environment map of a scene from a single LFOV LDRI.
Our model does not generate a border seam or any form of warping artefacts at the poles of the output HDRI. It can recreate a variety of both indoor and outdoor environments, outperforming the state of the art for both specialised scene types.
We compare our model against the current state of the art in LFOV illumination estimation using the evaluation method outlined by Weber et al. \cite{Weber2022}. We also compare against previous methods from a dataset kept updated by Dastjerdi et al. \cite{Dastjerdi2023}. Our model outperforms state-of-the-art models at removing ERP artefacts and illuminating objects with diffuse surfaces.  
To summarise our contributions are as follows:
\begin{itemize}
	\setlength\itemsep{0pt}
	\item First use of a purely Vision-Transformer network to approach the illumination estimation problem;
	\item The use of a Vision-Transformer network with global self attention creates more accurate diffuse lighting than current state-of-the-art methods;
	\item Incorporation of PanoSWIN attention modules and circular padding to better encode and generate ERPs by removing the warping artefacts that appear at the poles of panoramic images.
\end{itemize}

\section{Related Work}
\label{sec:review}

We first review the literature for methods that adapt neural networks to the ERP image format. We then review inverse tone-mapping techniques, a fundamental component that makes up the large majority of IBL illumination estimation papers and models. As our proposed network is based on ViTs, we briefly review transformer literature and focus on current Generative Adversarial Network (GAN) ViT architectures. Lastly, we investigate the current state-of-the-art illumination estimation papers and compare the benefits and compromises of each method.

\subsection{Equirectangular Panoramas and Neural Networks}
\label{ssec:ERP}

Various techniques have been proposed to overcome the artefacts created by the ERP format. Rotating the input image by 180$^\circ$, effectively turning it inside-out, has been used by \cite{Akimoto2019, Kim2021} to assist with panoramic outpainting by converting the problem from an unidirectional outpainting task to a part bidirectional inpainting task. This also helps reduce the seam generated at the sides of the panorama. To more reliably reduce the border and homogenise generation at the sides of the ERP, Akimoto et al. \cite{Akimoto2022} proposed a circular inference method for their transformer, removing the border seam but increasing inference time. Other methods have used circular padding either before and after or throughout the network \cite{Hara2020, Hilliard2023, Wang2023} allowing for homogenous generation and continuity at the sides of the ERP.
Several approaches have been adopted to overcome the warping at the top and bottom of the ERP. Conversion from ERP to cube map is one approach to remove warping, as used by \cite{Han2020, Chou2020, Feng2022, Gond2023}. However, as noted by \cite{Feng2022, Gond2023} additional requirements are needed to account for the seams generated at the intersections of the cube's faces. Feng \etal \cite{Feng2022} also observe that while the cube map format does well with local details it does not perform as well at capturing global context.
HEALPix \cite{Gorski}, a framework to give equal area weighting to spherical images, has been adopted by \cite{Chou2020, Yun} in their patch-embedding method to spherically encode the input image shape to a ViT. PAVER \cite{Yun} also uses a form of deform patch-embedding for their transformer architecture. Although these methods have been show to improve ViTs above baseline for the ERP format we found they do not remove border and warping artefacts in our tests.
A few papers have adapted loss functions to work with the ERP format. EnvMapNet \cite{Somanath2021} uses a spherical warping on the L1 loss and Akimoto et al. \cite{Akimoto2022} instead use the same spherical warping but on the Perceptual Loss. Another approach is to train the discriminator to identify ERP artefacts, by rotating the output ERP by 180$^\circ$ the border seam can then be detected by the discriminator. This has been implemented by \cite{Dastjerdi2022, Dastjerdi2023}.
Other approaches have aimed to change the architecture of neural networks to better handle ERP. Su and Grauman \cite{Su2022} develop a Spherical Convolution that adapts traditional CNNs to work with ERP. PanoSWIN \cite{Ling2023}, which outperformed Spherical Convolutions in benchmark metrics, uses a ViT architecture and observes that attention windows at the poles of the ERP should be considered connected. This changes the shifted window attention layout and implements a pitch attention module to further account for the ERP warping at the poles.

\subsection{Inverse Tone Mapping}
\label{ssec:ITM}

Inverse tone mapping is the process of converting an LDRI to an HDRI. It is commonly used in illumination estimation methods as HDR is an accurate way of capturing the lighting through an image. IBL illumination estimation methods perform inverse tone mapping as part of the extrapolation, however, some methods \cite{Gkitsas2020, Chen2022} have used a separate network to do the LDR to HDR conversion so that the main network can focus on field of view extrapolation.

Neural networks tend to be trained on LDRI with a range of [-1, 1] which works with the activation functions used, such as tanh activation with an output range of [-1, 1]. HDRIs have a much larger range (for example [0, 100000]) to capture the difference in light. A method for handling the range without biasing the loss functions to the larger values needs to be implemented. Gardner et al. \cite{Gardner2017a} separate the lighting into LDR and light log intensity. Other methods \cite{Zhang2017, Garon2019} use gamma correction, with $\alpha=1/30$ and $\gamma=2.2$ to map the HDR values visible in an LDRI to the same scale as an LDR. DeepLight \cite{Legendre2019} uses the natural log space to represent the illumination. Similarly, Li et al. \cite{Li2019} use the log(x+1) space. Hilliard et al. \cite{Hilliard2023} use a gamma compression with $\alpha=1$ and $\gamma=6.6$ so that the HDR is compressed to [0, 2] so that it can be better used with a tanh activation function, whilst increasing the range of the LDR values and decreasing the HDR values.

\subsection{Vision Transformers}
\label{ssec:ViT}

The transformer architecture was created to improve natural language processing by focusing on the relationships between each word and its surrounding words in a sentence. The Self-Attention and Feed Forward neural networks were later applied to the field of image processing with the ViT \cite{Dosovitskiy2020}, which used global self-attention to learn the relationship between each section (called a window) of the image and the image as a whole. Liu et al. proposed the SWIN Transformer, which used a hierarchical design with shifted attention windows and provided connections between them, allowing the network to use various scales and image sizes. ViTs were initially used for image classification, object detection and semantic segmentation but were later adapted into the GAN architecture for use in image generation and manipulation tasks. VQGAN \cite{Esser2021} used a CNN-based vector quantised variational autoencoder to train a ViT to learn a codebook of context-rich visual parts. RFormer \cite{Deng2022} combined ViTs with the U-Net style of architecture, using a ViT for the encoder and decoder, with shortcut connections between them, for image restoration. Gao et al. \cite{Gao2023} developed a similar U-Net style ViT called U-Transformer for image outpainting. This method used masked shortcut connections to connect only the known regions of the input image to the decoder layers.

\subsection{Illumination Estimation}
\label{ssec:IllEst}

Modern methods for illumination estimation extrapolated from an LFOV LDRI use deep learning to infer the lighting conditions. The format with which the lighting conditions are represented can be put into two distinct categories: regression based methods such as SH \cite{Cheng2018, Gkitsas2020, Garon2019} and SG \cite{Gardner2019, Li2019, Zhan2021b, Bai2023a} which use a basis function to reduce the lighting to a set of coefficients, and, as more recent methods \cite{Legendre2019, Somanath2021, Weber2022, Wang2022, Dastjerdi2023, Hilliard2023} are opting to use, an IBL approach by generating an ERP HDRI known as an environment map. Regression-based methods have the advantage of using less memory and are more efficient at rendering. However, IBL methods have been favoured because of the key advantage that they can be used to light mirror reflective surfaces due to their ability to represent high-frequency textures. The recent developments in IBL methods have aimed to create higher resolution and more plausible images that can generalise to a variety of indoor and outdoor scenes, whilst retaining accurate light positions and colour for lighting purposes. 
The illumination estimation methods that make adaptions for ERP are \cite{Somanath2021, Dastjerdi2023, Akimoto2022, Dastjerdi2022}. These methods have only focused on changing the loss functions rather than the architecture of the network. Consequently, they are unable to generate the poles of the ERP without noticeable artefacts.

\section{Methodology}
\label{sec:method}

\begin{figure*}[tb]
	\begin{subfigure}[t]{1.0\linewidth}
		\centering
		\includegraphics[width=\textwidth]{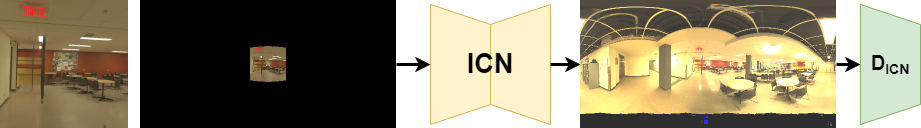}
	\end{subfigure}
	  \vspace{0.3ex} 

	 \rule[0.5ex]{\linewidth}{.4pt} 

	\begin{subfigure}[t]{1.0\linewidth}
		\centering
		\includegraphics[width=\textwidth]{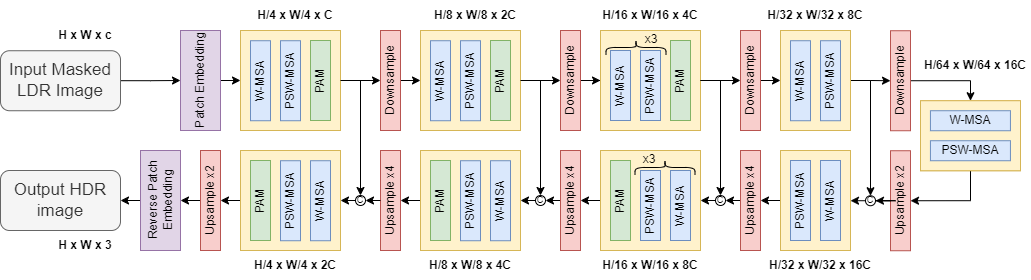}
	\end{subfigure}
\caption{Summary of the proposed model. \textit{Top:} The overall flow of the model. The generator (ICN) uses the masked LDR ERP as input to generate the HDR ERP environment map. This is trained as a GAN by the discriminator ($D_{ICN}$). \textit{Bottom:} The 360U-Former architecture used by the ICN. The PanoSWIN attention blocks W-MSA, PSW-MSA and PAM are desribed in \cref{ssec:Network} and \cref{fig:panoswin}.}
\label{fig:model}
\end{figure*}

We present 360U-Former, a U-Net style ViT-based architecture trained as a GAN to generate HDR 360$^\circ$ ERP images from LFOV LDR images that can be applied to both indoor and outdoor scenes. By using PanoSWIN attention layers our model generates ERP images without warping at the poles and a seam at the sides of the image. The architecture for the discriminator is based on RALSGAN \cite{Jolicoeur-Martineau2018}. The pipeline for the whole model is featured in \cref{fig:model}.

\subsection{Network Architecture}
\label{ssec:Network}

\begin{figure*}[t]
	\begin{subfigure}[b]{0.32\linewidth}
		\centering
		\centerline{\includegraphics[width=\textwidth]{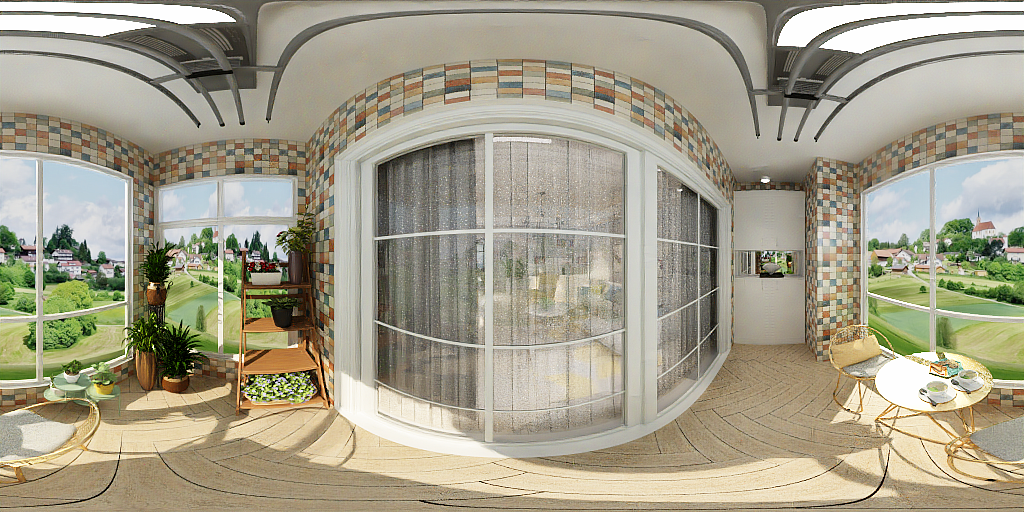}}
		\caption{W-MSA}\medskip
        \label{fig:panoswin-a}
	\end{subfigure}
	\hfill
	\begin{subfigure}[b]{0.32\linewidth}
		\centering
		\centerline{\includegraphics[width=\textwidth]{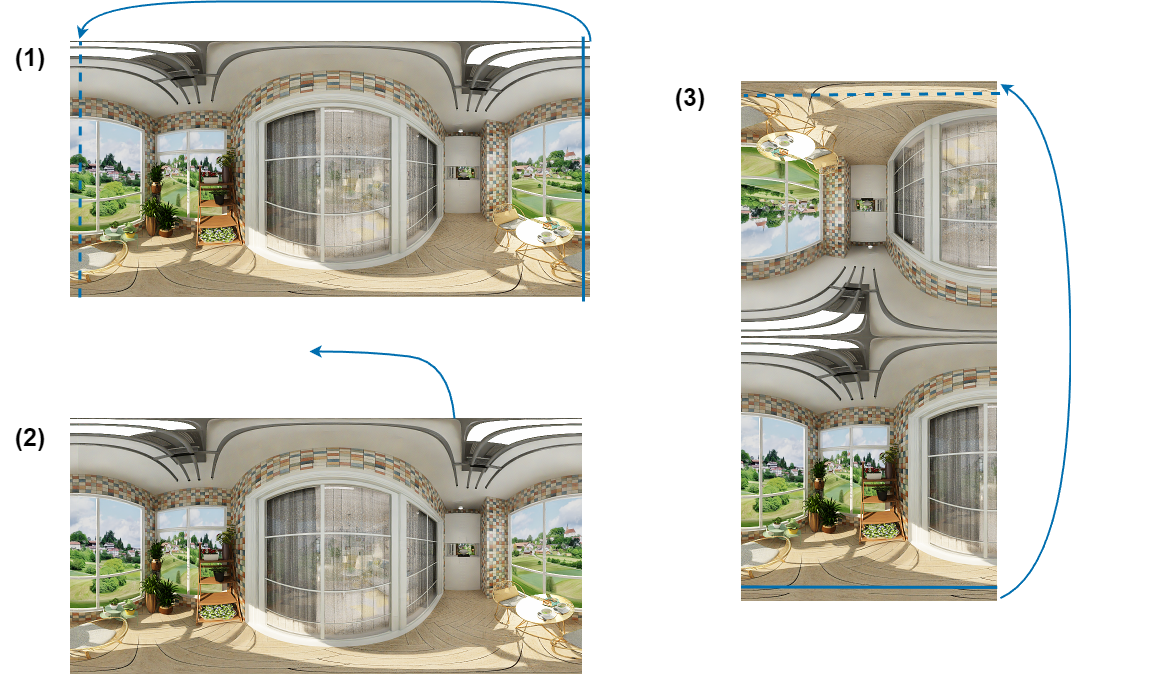}}
		\caption{PSW-MSA}\medskip
        \label{fig:panoswin-b}
	\end{subfigure}
	\hfill
	\begin{subfigure}[b]{0.32\linewidth}
		\centering
		\centerline{\includegraphics[width=\textwidth]{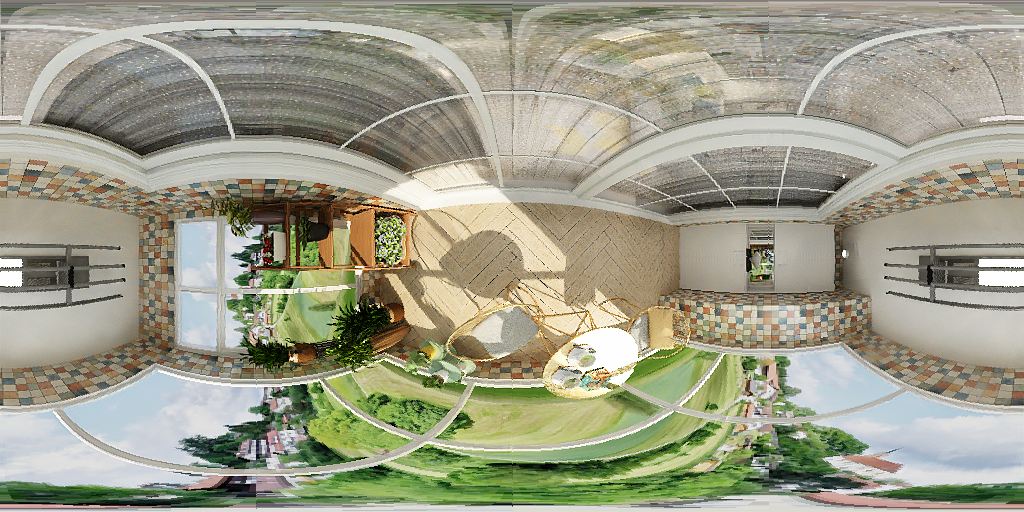}}
		\caption{PAM}\medskip
        \label{fig:panoswin-c}
	\end{subfigure}
\caption{The ERP rotations that are used as input by each of the three attention layers.}
\label{fig:panoswin}
\end{figure*}

The input to the network is an LFOV LDR image converted to ERP format. The output of the model is an HDR ERP image. 
The 360U-Former, \cref{fig:model}, can be split into three distinct parts: the encoder, the bottleneck and the decoder.

The encoder uses 4 PanoSWIN \cite{Ling2023} blocks with layer sizes of 3, 3, 7  and 2. The PanoSWIN block consists of a standard window multi-head self-attention layer (W-MSA), \cref{fig:panoswin-a} shows the input, followed by a panoramically warped shifted window multi-head self-attention layer (PSW-MSA), \cref{fig:panoswin-b} shows the rotations done and the final input to the layer, that ensures that the regions along the top or bottom of the panorama are considered adjacent rather than distant. The final layer of each block, except the last block, is a Pitch Attention Module (PAM). The PAM performs cross-attention on the windows from the default orientation and the associated windows from the image pitched and rotated by 90$^\circ$, \cref{fig:panoswin-c}. This method allows the network to learn the spatial distortion at the poles of the ERP. For the bottleneck, we use a block of two PanoSWIN layers without the PAM. We found that using the PAM in the bottleneck, the last block of the encoder and the first block of the decoder would lead to artefacts at the sides of the generated image. This is likely due to the size of the tensor at these blocks, 8 by 16 and 4 by 8, and the way they are interpolated that affects the edge of the output image.

The decoder is similar to the encoder but uses upsampling instead of down-sampling to increase the resolution and reduce the channel size. We incorporate shortcut connections from the encoder by further upsampling the channels of the tensor from the previous block and concatenating them with the associated output from the encoder. When reversing the patch embedding to get the output image we incorporate circular padding on either side of the image and reflection padding at the top and bottom instead of the default padding used by convolutional layers. This further ensures that no border artefacts are produced.

\subsection{Loss Functions}
\label{ssec:loss}

To train our network we use three loss functions. We choose L1 loss for pixel-wise accuracy. To measure semantic similarity to the ground truth we use the perceptual loss \cite{Zhang2018} $L_{perc}$.
\begin{align}
L_{perc}=\sum_l\frac{1}{\sum_{u,v}}\sum_{u,v} \llbracket \left(y^l_{uv}-x^l_{uv}\right)\rrbracket,
\end{align}
where $u$ and $v$ are the positions on the feature (size $H_l \times W_l$) in the $l$th layer of the VGG feature extractor.

We leverage the RalSGAN \cite{Jolicoeur-Martineau2018} to train our GAN because we found it to generate higher-quality images with fewer artefacts.
\begin{align}
\begin{aligned}
L_{adv} = \mathbb{E}\left[\log D(I_H)\right]+\mathbb{E}\left[\log \left[1-D(G_{ICN}(\overline{I}_L))\right]\right].
\end{aligned}
\end{align}
The ICN overall objective with weighted loss functions is
\begin{align}
L_{G_{ICN}}=\lambda_{L1}L_{L1}+\lambda_{perc}L_{perc}+\lambda_{adv}L_{adv},
\end{align}
where $\lambda$ represents the loss weight for each loss. We found the optimal loss weights to be $\lambda_{L1}=5$, $\lambda_{perc}=5$ and $\lambda_{adv}=0.2$ by testing a range of values and comparing the performance of the metrics and appearance of the generated output.

\subsection{Implementation Details}
\label{ssec:implement}

We train our network on both indoor and outdoor data. For the indoor dataset, we use the Structured3D \cite{Zheng2019} synthetic dataset which contains 21,843 photo-realistic panoramas. For the outdoor dataset, we use the 360 Sun Positions \cite{Chang2018} dataset which contains 19,093 street view images of both urban roads and rural environments.

To create the HDR ground truth pairs for these LDR datasets, we convert to HDR using an LDR-to-HDR network. This network is based on the network design from Bolduc \etal \cite{Bolduc2023} without the exposure and illuminance branch using the ERP LDRI as input and the ERP HDRI as output. It is trained on the Laval Photometric Indoor HDR dataset \cite{Bolduc2023}, Laval Outdoor HDR dataset \cite{Hold-Geoffroy2019} and the dataset from Cheng \etal \cite{Cheng2018}. Due to the difference in calibration between each dataset, we base the values on the average mean and median values from the Laval Photometric dataset scaled by a factor 0.01. This scaling factor is based on the factor needed to create a plausible render when using the Cycles rendering engine in Blender. Using this method we used a scaling factor of 340 for the Laval HDR Outdoor dataset and that the other dataset did not need to be scaled.

The network is trained on an image resolution of 256 by 512. We augment both datasets by horizontally rotating each panorama 8 times randomly between 20$^\circ$ and 340$^\circ$ at intervals of 40$^\circ$ with a 20\% chance of being vertically flipped. After augmentation our dataset consists of 368,370 images pairs which we split into a train/test ratio of 99:1. We ensure that all augmented versions of each pair exist only in the train or the test subset to prevent over-fitting. Both networks are trained with a range of LFOV sizes $\nobreak \{40^\circ, 60^\circ, 90^\circ, 120^\circ\}$ for the masked input. The mask size is randomly chosen for each input image.
The network is trained for 50 epochs on an A100 GPU at 5.5 hours per epoch. There are a total of 220 million parameters in the generator network. We use the ADAM optimiser with betas 0 and 0.9, weight decay of 0.0001 and learning rate of 0.0001 for the generator and 0.0004 for the discriminator.

\section{Evaluation}
\label{sec:evaluation}

We evaluate our model against the latest state-of-the-art illumination estimation methods using the protocol first outlined by Weber et al. \cite{Weber2022} and used again by EverLight \cite{Dastjerdi2023}. We compare quantitative results for indoor and outdoor methods separately as some previous methods focused on one or the other. This also helps to highlight the performance in different domains. To highlight the ability of our network to adapt to the ERP format, we rotate the outputs to show the border seams and the quality of the poles generated. We also conduct an ablation study to compare the PanoSWIN against the SWIN attention blocks. The results of the diffuse render tests have been generously contributed to the community by Dastjerdi \etal \cite{Dastjerdi2023} and can be found on the project website of EverLight. \footnote{\url{https://lvsn.github.io/everlight/}}

\subsection{Indoor Images}
\label{ssec:indoor}

\begin{table}[t]
\caption{Indoor and outdoor environment quantitative comparison with various illumination estimation methods. The metrics si-RMSE, RMSE, RGB ang. and PSNR are evaluated by rendering a diffuse scene and computing the differences between the tonemapped renders. The FID score is calculated on the generated environment maps. The best scores for each metric and environment are highlighted in {\bf bold} and the second best \underline{underlined}.}
\label{T_Indoor}
\centering
{
\scriptsize
\begin{tabular}{|l|l|l|l|l|l|}
\hline
Method & si-RMSE$\downarrow$ & RMSE$\downarrow$ & RGB ang.$\downarrow$ & PSNR$\uparrow$ & FID$\downarrow$\\
\hline\hline
\multicolumn{6}{|l|}{INDOOR METHODS}\\
\hline
Ours&{\bf 0.033}&{\bf 0.110}&6.11$^\circ$&11.68&119.91\\
\hline
Hilliard'23\cite{Hilliard2023}&0.112&0.300&6.50$^\circ$&10.05&158.60\\
\hline
EverLight\cite{Dastjerdi2023}&0.087&0.239&5.75$^\circ$&10.04&\underline{65.50}\\
\hline
Weber'22\cite{Weber2022}&\underline{0.079}&{\underline{0.196}}&\underline{4.08$^\circ$}&{\bf 12.95}&130.13\\
\hline
StyleLight\cite{Wang2022}&0.130&0.261&7.05$^\circ$&\underline{12.85}&121.60\\
\hline
Gardner'19(1)\cite{Gardner2019}&0.099&0.229&4.42$^\circ$&12.21&410.12\\
\hline
Gardner'19(3)\cite{Gardner2019}&0.105&0.507&4.59$^\circ$&10.90&386.43\\
\hline
Gardner'17\cite{Gardner2017a}&0.123&0.628&8.29$^\circ$&10.22&253.40\\
\hline
Garon'19\cite{Garon2019}&0.096&0.255&8.06$^\circ$&9.73&324.51\\
\hline
Lighthouse&0.121&0.254&4.56$^\circ$&9.81&174.52\\
\hline
EMLight\cite{Zhan2021b}&0.099&0.232&{\bf 3.99$^\circ$}&10.34&135.97\\
\hline
EnvmapNet\cite{Somanath2021}&0.097&0.286&7.67$^\circ$&11.74&221.85 \\
\hline
ImmerseGAN\cite{Dastjerdi2022}&0.091&0.215&7.89$^\circ$&10.87&{\bf 55.46}\\
\hline\hline
\multicolumn{6}{|l|}{OUTDOOR METHODS}\\
\hline
Ours&{\bf 0.049}&{\bf 0.161}&{\bf 4.00$^\circ$}&{\bf 13.27}&102.63\\
\hline
EverLight\cite{Dastjerdi2023}&\underline{0.162}&0.385&\underline{8.30$^\circ$}&\underline{11.01}&\underline{61.49}\\
\hline
ImmerseGAN\cite{Dastjerdi2022}&0.175&\underline{0.341}&9.56$^\circ$&10.91&{\bf 34.43}\\
\hline
Zhang '19\cite{Zhang2019a} &0.225&1.058&11.80$^\circ$&10.91&449.49\\
\hline
\end{tabular}
}
\vskip .25cm

\end{table}

\begin{figure*}
\begin{center}
\setlength\tabcolsep{0pt}
\renewcommand{\arraystretch}{0.25}
\begin{tabularx}{\linewidth}{lXXXXX}

\rotatebox{90}{\scriptsize Input} &\centering \includegraphics[width=0.12\textwidth]{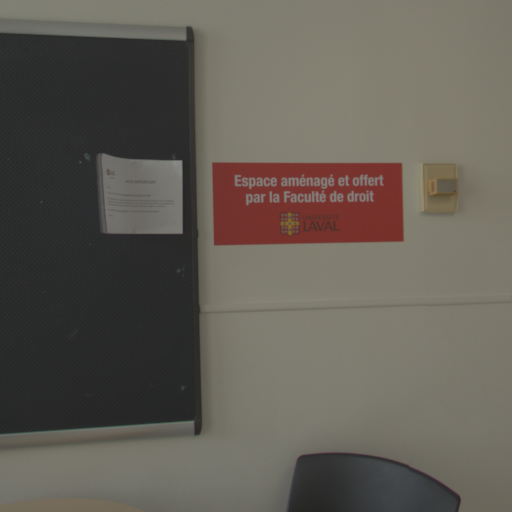} &\centering \includegraphics[width=0.12\textwidth]{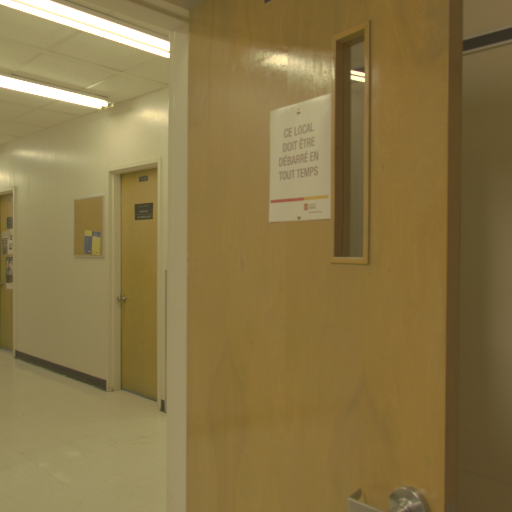} &\centering \includegraphics[width=0.12\textwidth]{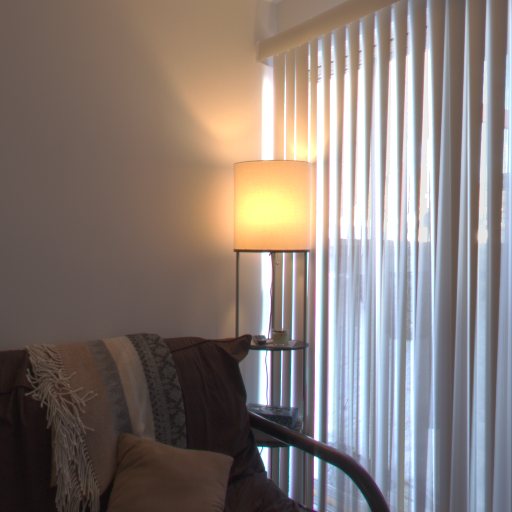} &\centering \includegraphics[width=0.12\textwidth]{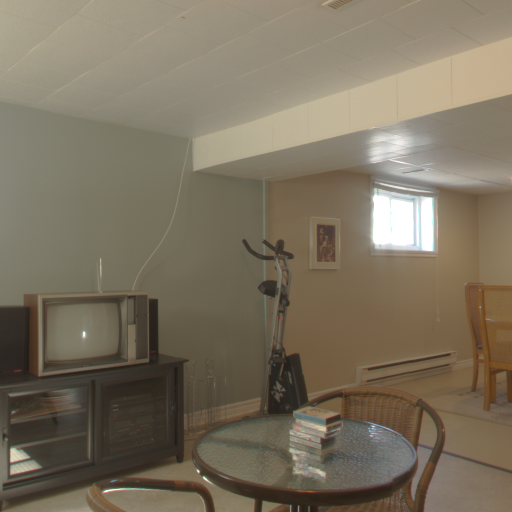} &\multicolumn{1}{c}{\includegraphics[width=0.12\textwidth]{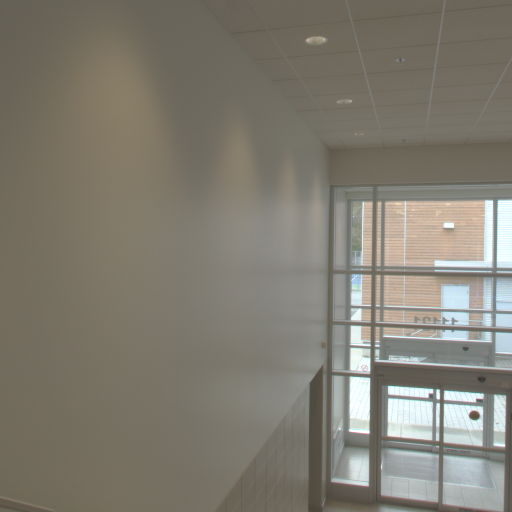}}\\

\multirow{2}*{\rotatebox{90}{\scriptsize Ground Truth           }}&&&&\\

&\includegraphics[width=0.19\textwidth]{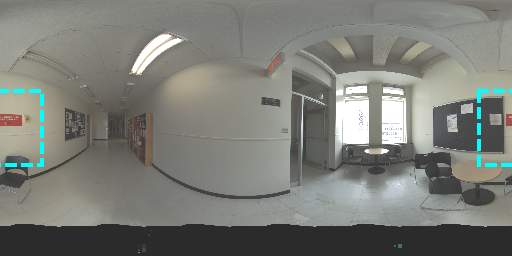}
&\includegraphics[width=0.19\textwidth]{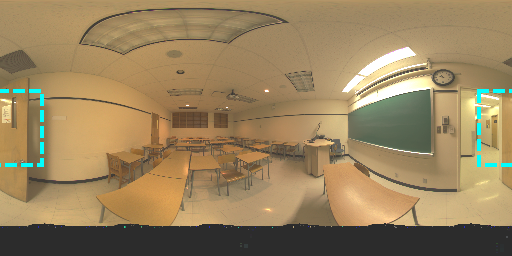} &\includegraphics[width=0.19\textwidth]{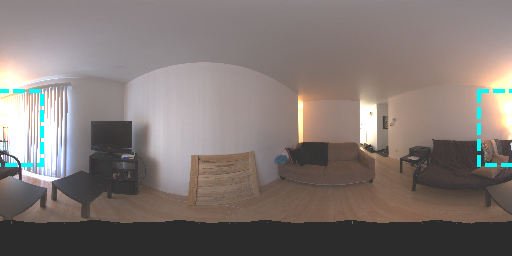} &\includegraphics[width=0.19\textwidth]{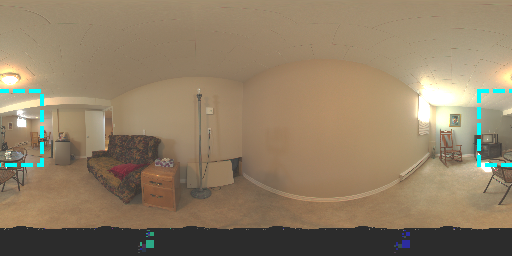} &\includegraphics[width=0.19\textwidth]{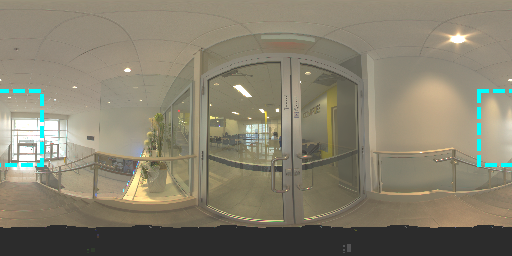}\\

\multirow{2}*{\rotatebox{90}{\scriptsize Gardner '17 \cite{Gardner2017a}} 
 }&\includegraphics[width=0.19\textwidth]{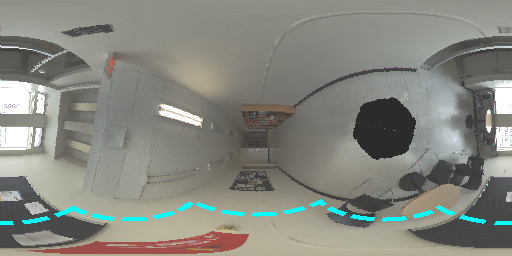}
&\includegraphics[width=0.19\textwidth]{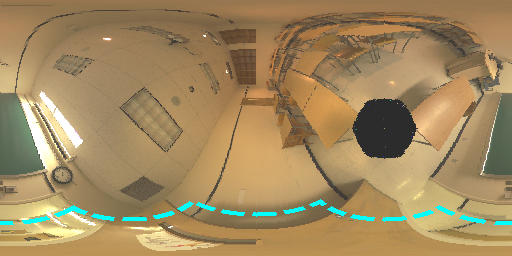} &\includegraphics[width=0.19\textwidth]{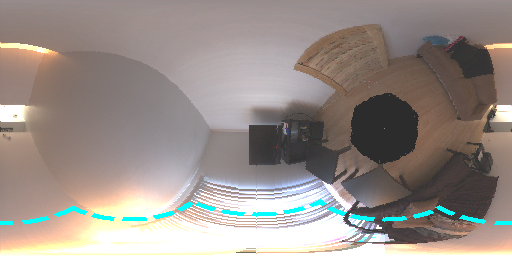} &\includegraphics[width=0.19\textwidth]{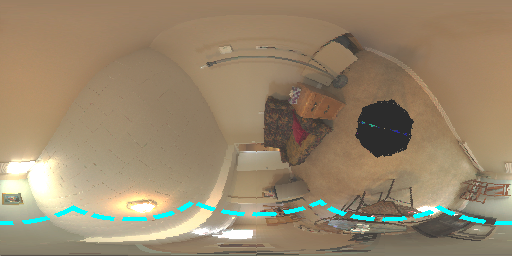} &\includegraphics[width=0.19\textwidth]{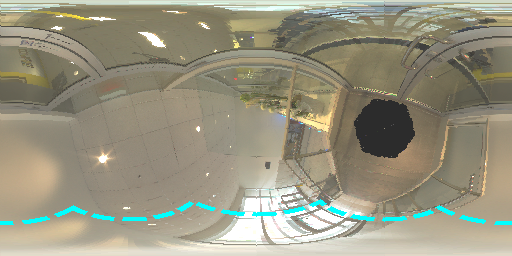}\\

&\includegraphics[width=0.19\textwidth]{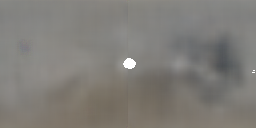}
&\includegraphics[width=0.19\textwidth]{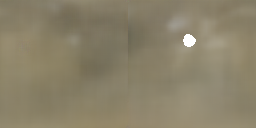} &\includegraphics[width=0.19\textwidth]{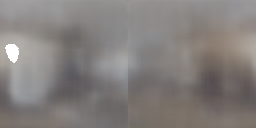} &\includegraphics[width=0.19\textwidth]{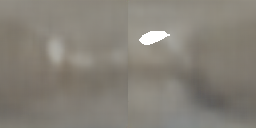} &\includegraphics[width=0.19\textwidth]{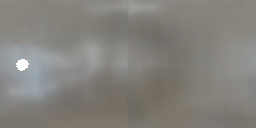}\\

\multirow{2}*{\rotatebox{90}{\scriptsize Weber '22 \cite{Weber2022} }} &\includegraphics[width=0.19\textwidth]{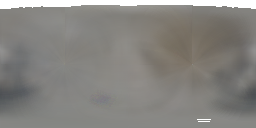}
&\includegraphics[width=0.19\textwidth]{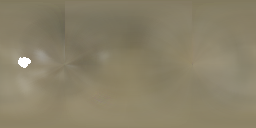} &\includegraphics[width=0.19\textwidth]{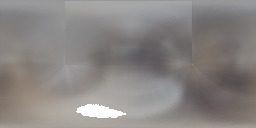} &\includegraphics[width=0.19\textwidth]{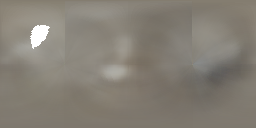} &\includegraphics[width=0.19\textwidth]{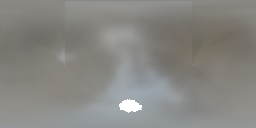}\\

&\includegraphics[width=0.19\textwidth]{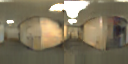}
&\includegraphics[width=0.19\textwidth]{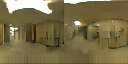} &\includegraphics[width=0.19\textwidth]{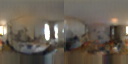} &\includegraphics[width=0.19\textwidth]{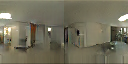} &\includegraphics[width=0.19\textwidth]{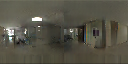}\\

\multirow{2}*{\rotatebox{90}{\scriptsize StyleLight \cite{Wang2022}  }} &\includegraphics[width=0.19\textwidth]{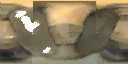} &\includegraphics[width=0.19\textwidth]{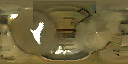} &\includegraphics[width=0.19\textwidth]{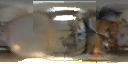} &\includegraphics[width=0.19\textwidth]{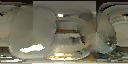} &\includegraphics[width=0.19\textwidth]{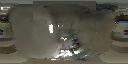}\\

&\includegraphics[width=0.19\textwidth]{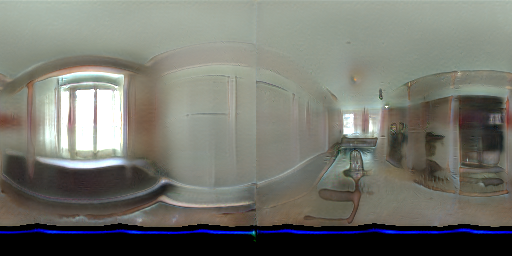} &\includegraphics[width=0.19\textwidth]{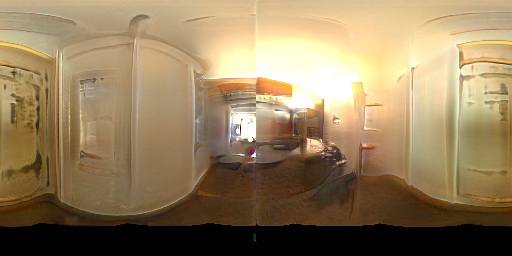} &\includegraphics[width=0.19\textwidth]{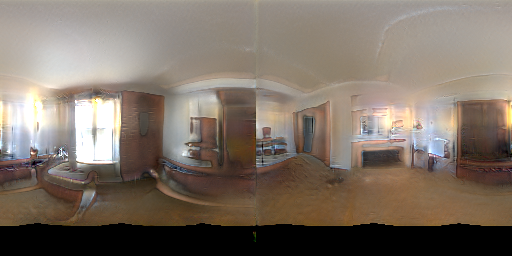} &\includegraphics[width=0.19\textwidth]{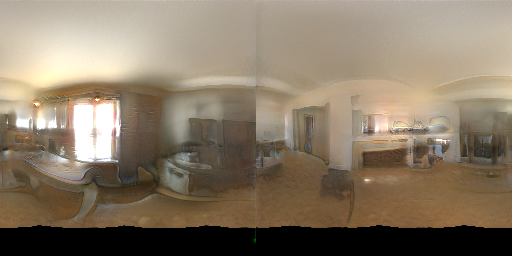} &\includegraphics[width=0.19\textwidth]{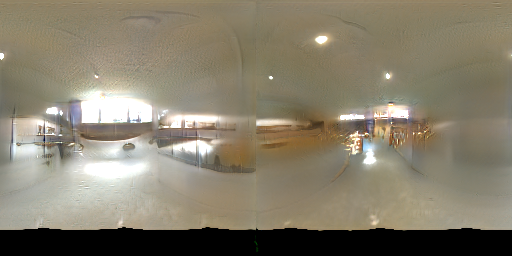}\\




\multirow{2}*{\rotatebox{90}{\scriptsize EverLight \cite{Dastjerdi2023}}} &\includegraphics[width=0.19\textwidth]{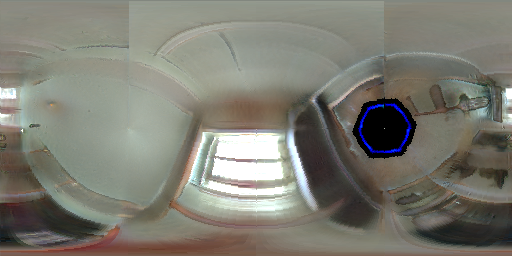} &\includegraphics[width=0.19\textwidth]{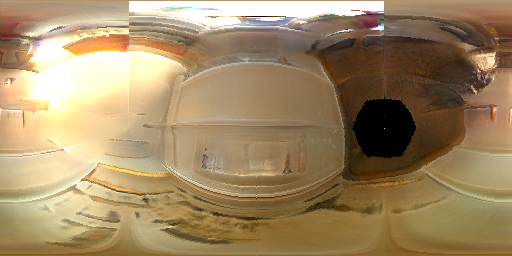} &\includegraphics[width=0.19\textwidth]{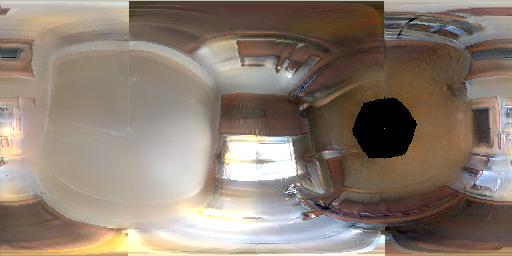} &\includegraphics[width=0.19\textwidth]{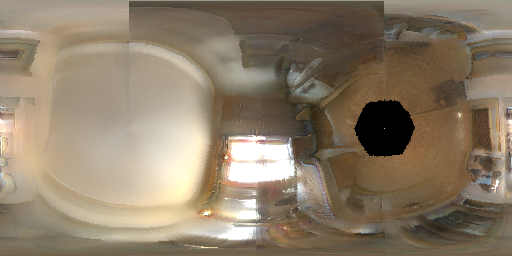} &\includegraphics[width=0.19\textwidth]{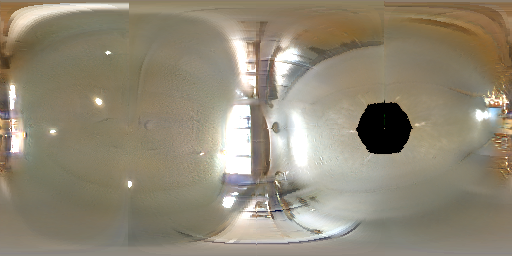}\\

&\includegraphics[width=0.19\textwidth]{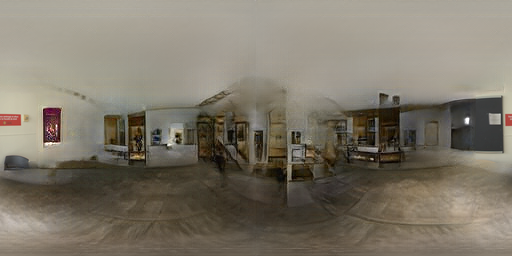} &\includegraphics[width=0.19\textwidth]{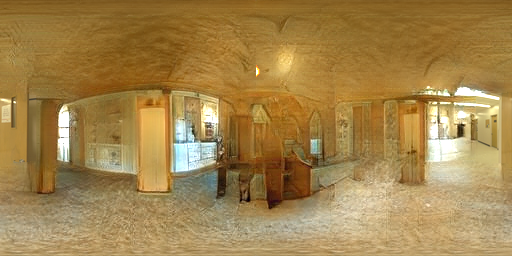} &\includegraphics[width=0.19\textwidth]{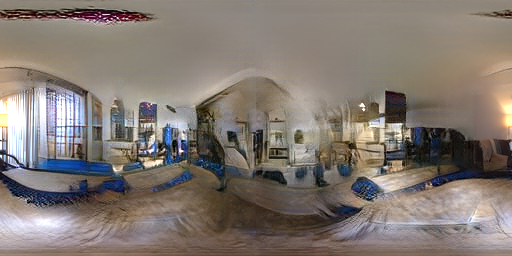} &\includegraphics[width=0.19\textwidth]{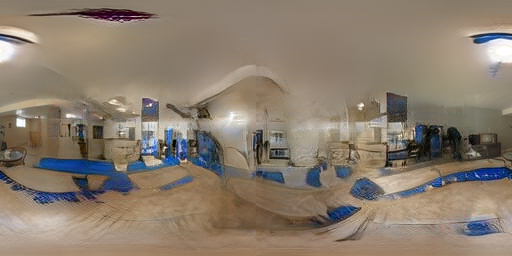} &\includegraphics[width=0.19\textwidth]{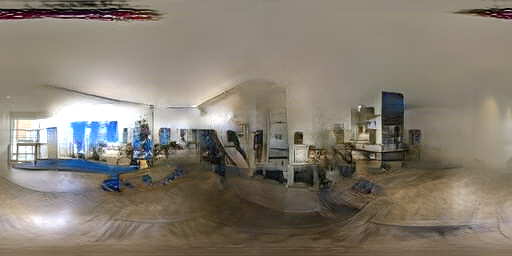}\\

&\includegraphics[width=0.19\textwidth]{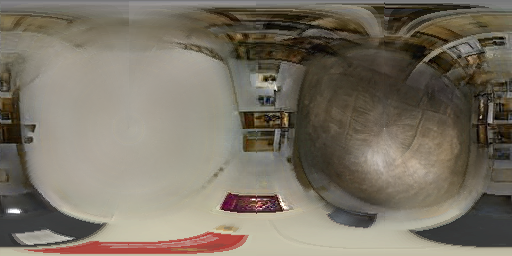}
&\includegraphics[width=0.19\textwidth]{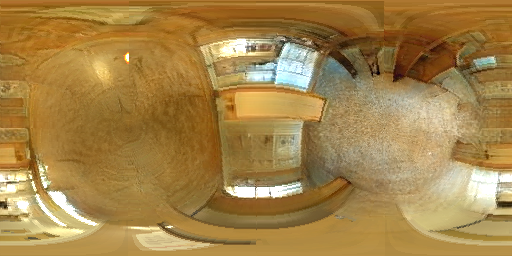} &\includegraphics[width=0.19\textwidth]{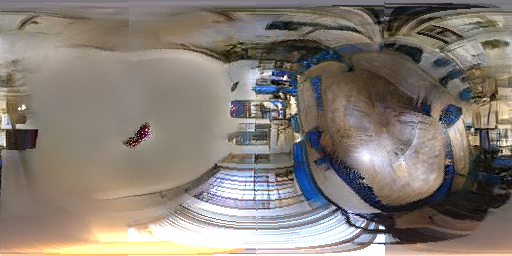} &\includegraphics[width=0.19\textwidth]{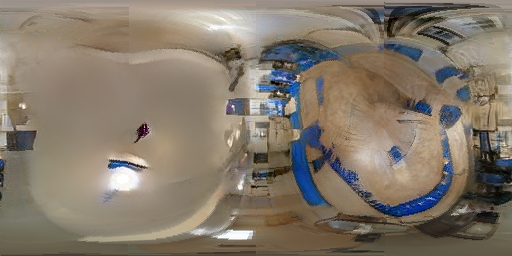} &\includegraphics[width=0.19\textwidth]{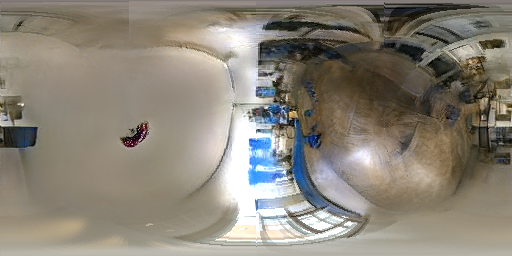}\\

\rotatebox{90}{\scriptsize Ours} &\includegraphics[width=0.19\textwidth]{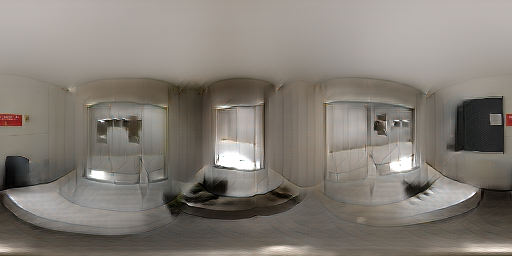} &\includegraphics[width=0.19\textwidth]{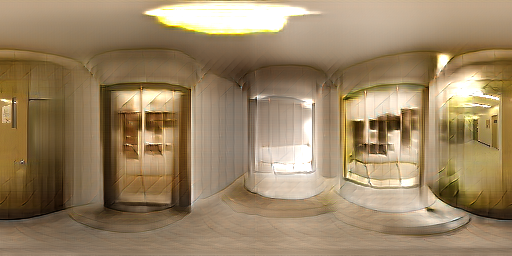} &\includegraphics[width=0.19\textwidth]{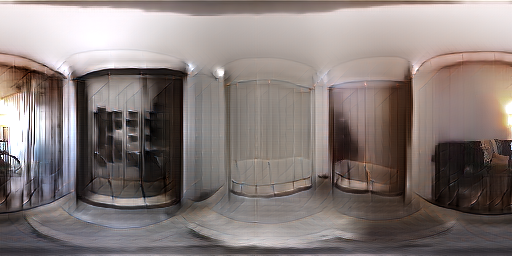} &\includegraphics[width=0.19\textwidth]{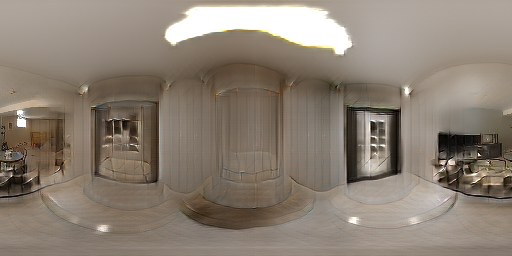} &\includegraphics[width=0.19\textwidth]{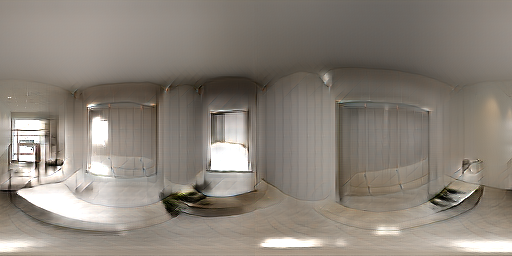}\\

&\includegraphics[width=0.19\textwidth]{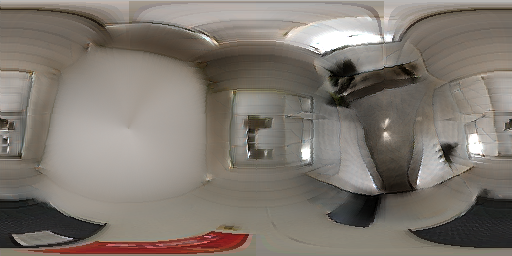} &\includegraphics[width=0.19\textwidth]{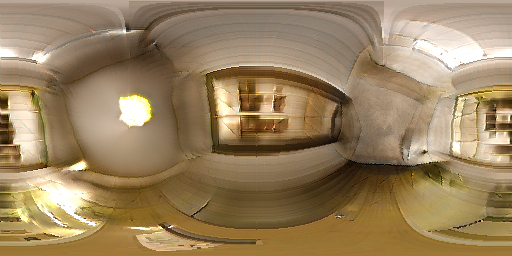} &\includegraphics[width=0.19\textwidth]{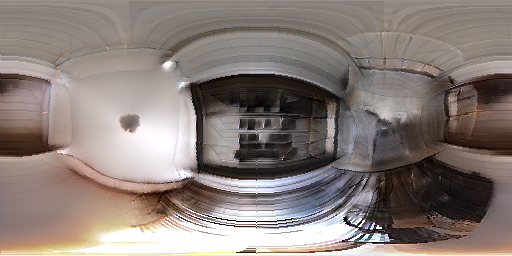} &\includegraphics[width=0.19\textwidth]{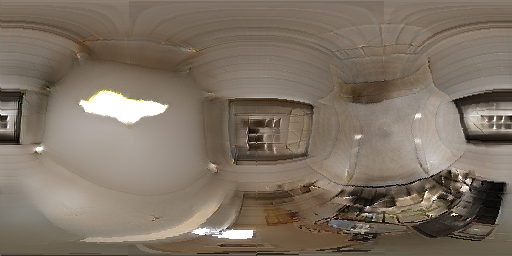} &\includegraphics[width=0.19\textwidth]{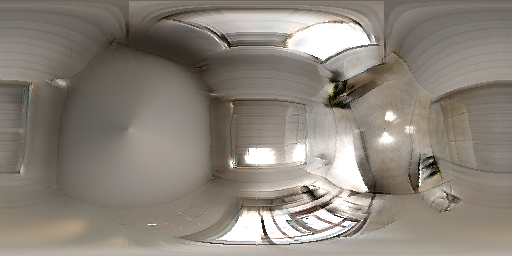}\\
\end{tabularx}
\end{center}
\caption{Indoor qualitative comparison of our generated ERPs in LDR with other methods. For each method and input LFOV image we show the LDR ERP rotated 180$^\circ$ to show any potential border seams and the LDR ERP rotated 90$^\circ$ by 90$^\circ$ to compare the generation at the poles of the ERP. We only include a selection of the methods from the quantitative comparison, the remaining methods can be found in the supplementary material. For the ground truth, we show the input to the network and include a dotted box around that area in the panorama.}
\label{fig:indoor_renders}
\end{figure*}

\subsubsection{Quantitative Results}
\label{sssec:indoor_quant}

We follow the evaluation protocol of Everlight \cite{Dastjerdi2023}, testing using a two-fold method. First, evaluating the ability of the method to produce accurate light positions, colours and intensities on a diffuse surface. This is conducted using the test split of the Laval HDR Indoor dataset \cite{Gardner2017a} and 10 extracted views of 50$^\circ$. The generated environment maps are rendered to light a scene with 9 diffuse spheres on a ground plane. We use RMSE, scale-invariant RMSE (siRMSE), RGB angular error and PSNR metrics to measure the diffuse lighting accuracy. Second, the plausibility of the generated HDRIs is measured with Fréchet Inception Distance (FID). We evaluate the FID score in the same way as \cite{Dastjerdi2023}, using additional datasets to remove the potential bias of using just one dataset. These are the 305 images from the Laval HDR Indoor test set \cite{Gardner2017a} and the 192 indoor images from \cite{Cheng2018}. However, as the 360Cities dataset is not publicly available we calculate the FID score for our work without these images. As with the rendered evaluation, we take 10 images with a field of view of 50$^\circ$ extracted from the ground truth panorama as input to the network.

Following the protocol, we evaluate against the following methods. Two versions of \cite{Gardner2019} are compared: the original (3) where 3 light sources are estimated, and a version (1) trained to predict a single parametric light. We also compare to Lighthouse \cite{Srinivasan2020}, which expects a stereo pair as input, instead a second image is generated with a small baseline using \cite{Wiles2020} (visual inspection confirmed this yields results comparable to the published work). For \cite{Garon2019}, the coordinates of the image centre for the object position are selected. For \cite{Somanath2021}, the proposed “Cluster ID loss”, and tonemapping are used with a pix2pixHD \cite{Wang2018} network architecture. We compare against EMLight \cite{Zhan2021b}, StyleLight \cite{Wang2022}, EverLight \cite{Dastjerdi2023}, \cite{Weber2022} and \cite{Hilliard2023}. Finally, \cite{Dastjerdi2022} as a state-of-the-art (LDR) LFOV extrapolation method.

The results for the Indoor quantitative comparison are shown in \cref{T_Indoor}. In terms of si-RMSE and RMSE, our method outperforms all other methods indicating that the colours are more accurate. In terms of light position accuracy, the methods that incorporate parametric or spherical Gaussian lighting as the main form of lighting representation perform better. Our method performs competitively in terms of PSNR and FID scores.

\subsubsection{Qualitative Results}
\label{sssec:indoor_qual}

\begin{figure}[t]
\begin{center}
\setlength\tabcolsep{0pt}
\renewcommand{\arraystretch}{0.25}
\begin{tabularx}{\linewidth}{lXXXXX}

\rotatebox{90}{Input} &\centering \includegraphics[width=0.13\textwidth]{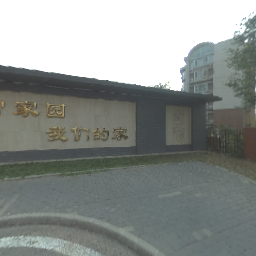} &\centering \includegraphics[width=0.13\textwidth]{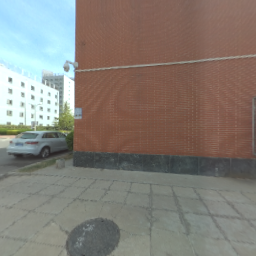} &\centering \includegraphics[width=0.13\textwidth]{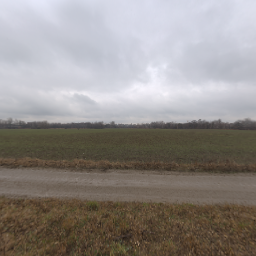} &\centering \includegraphics[width=0.13\textwidth]{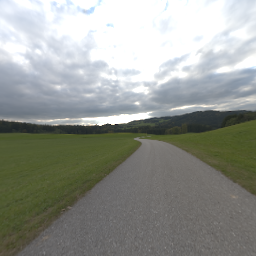} &\multicolumn{1}{c}{\includegraphics[width=0.13\textwidth]{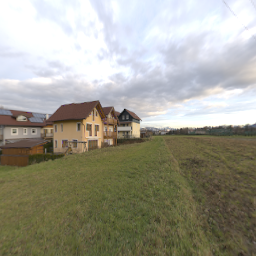}}\\

\multirow{2}*{\rotatebox{90}{Ground Truth  }}&&&&\\

&\includegraphics[width=0.19\textwidth]{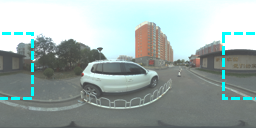}
&\includegraphics[width=0.19\textwidth]{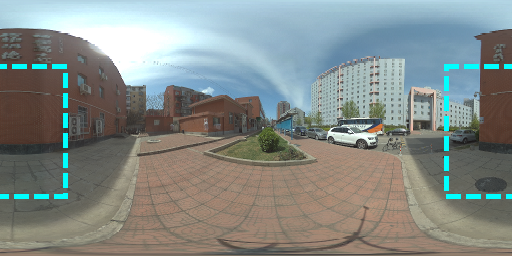} &\includegraphics[width=0.19\textwidth]{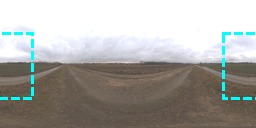} &\includegraphics[width=0.19\textwidth]{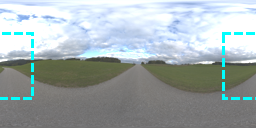} &\includegraphics[width=0.19\textwidth]{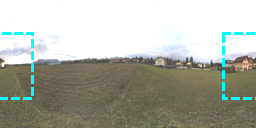}\\

\multirow{2}*{\rotatebox{90}{ImmerseGAN \cite{Dastjerdi2022}}} &\includegraphics[width=0.19\textwidth]{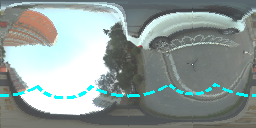}
&\includegraphics[width=0.19\textwidth]{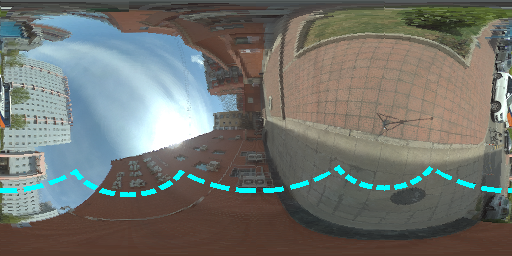} &\includegraphics[width=0.19\textwidth]{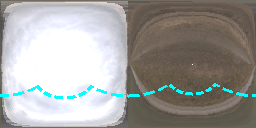} &\includegraphics[width=0.19\textwidth]{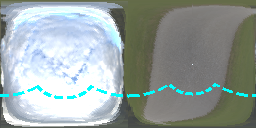} &\includegraphics[width=0.19\textwidth]{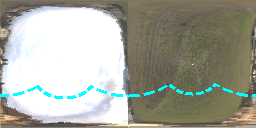}\\

&\includegraphics[width=0.19\textwidth]{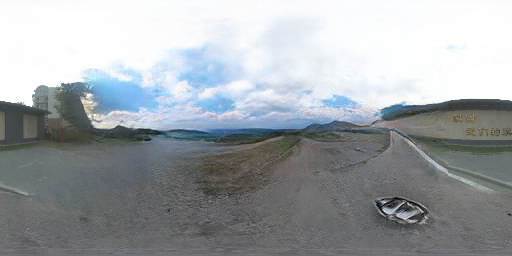}
&\includegraphics[width=0.19\textwidth]{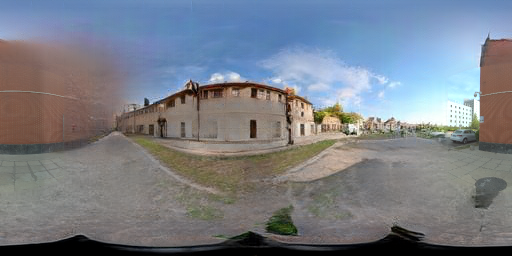} &\includegraphics[width=0.19\textwidth]{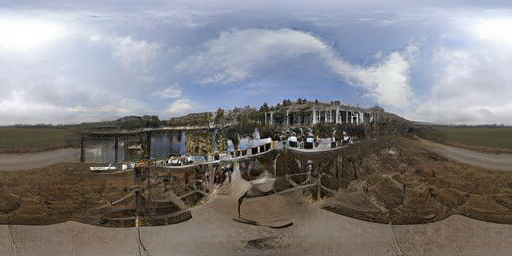} &\includegraphics[width=0.19\textwidth]{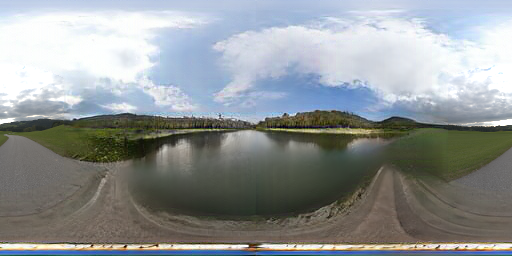} &\includegraphics[width=0.19\textwidth]{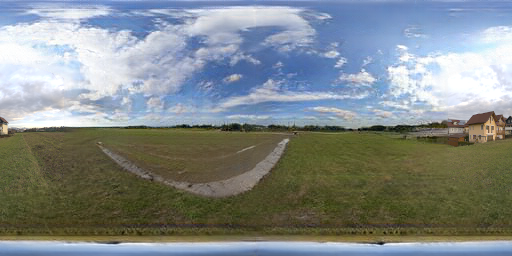}\\

\multirow{2}*{\rotatebox{90}{EverLight \cite{Dastjerdi2023}   }} &\includegraphics[width=0.19\textwidth]{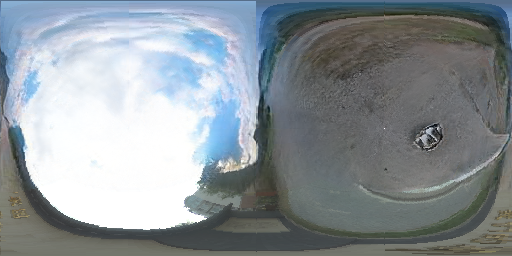}
&\includegraphics[width=0.19\textwidth]{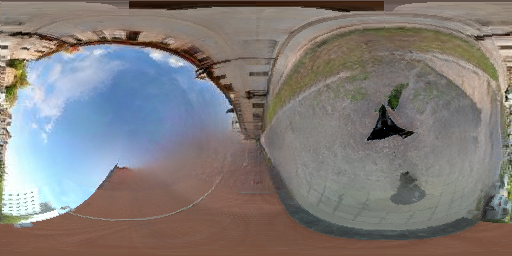} &\includegraphics[width=0.19\textwidth]{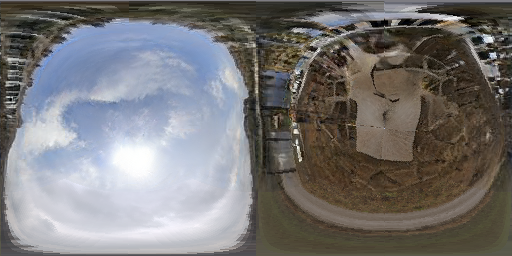} &\includegraphics[width=0.19\textwidth]{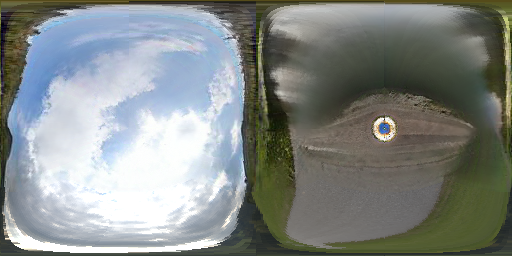} &\includegraphics[width=0.19\textwidth]{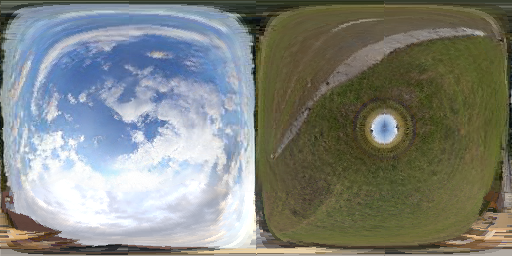}\\

&\includegraphics[width=0.19\textwidth]{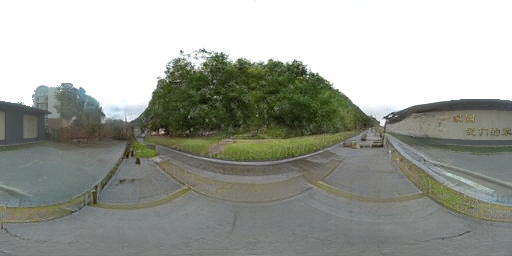}
&\includegraphics[width=0.19\textwidth]{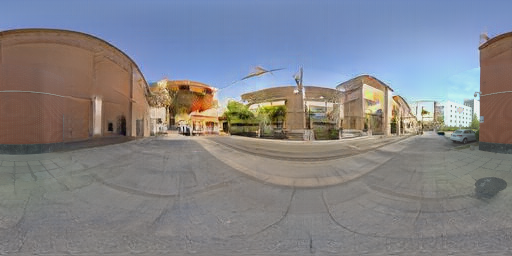} &\includegraphics[width=0.19\textwidth]{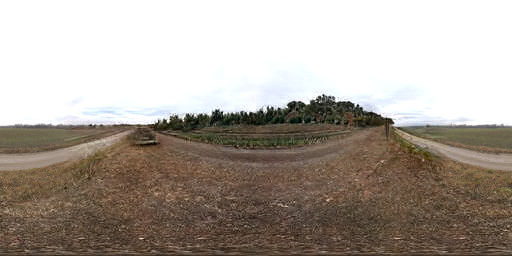} &\includegraphics[width=0.19\textwidth]{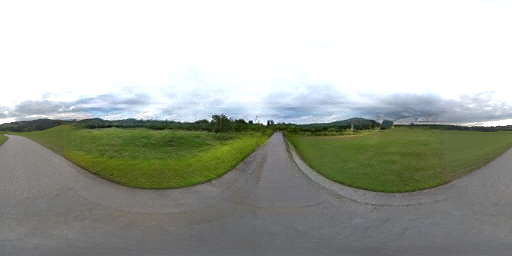} &\includegraphics[width=0.19\textwidth]{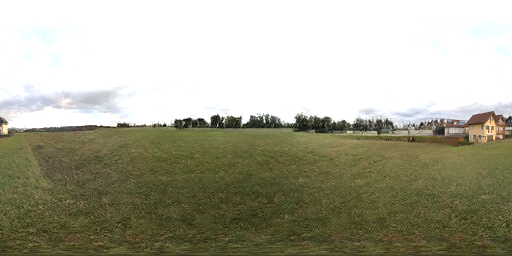}\\

&\includegraphics[width=0.19\textwidth]{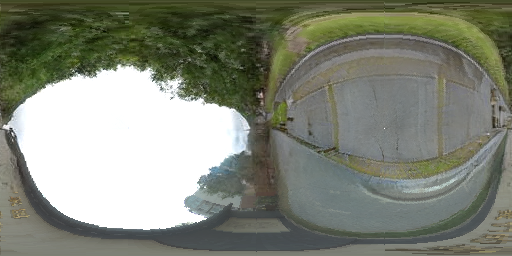}
&\includegraphics[width=0.19\textwidth]{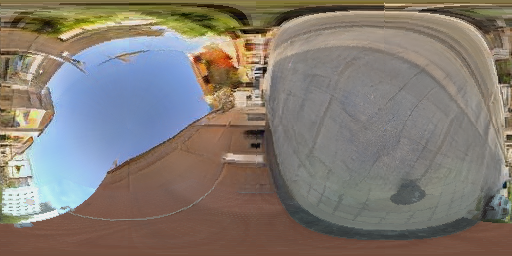} &\includegraphics[width=0.19\textwidth]{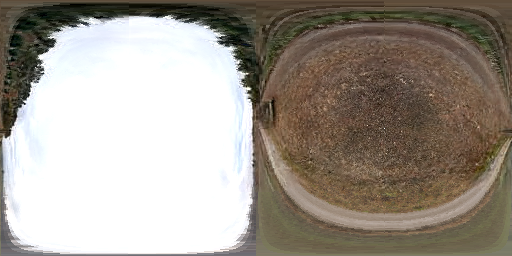} &\includegraphics[width=0.19\textwidth]{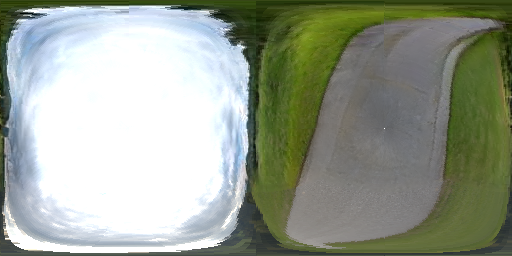} &\includegraphics[width=0.19\textwidth]{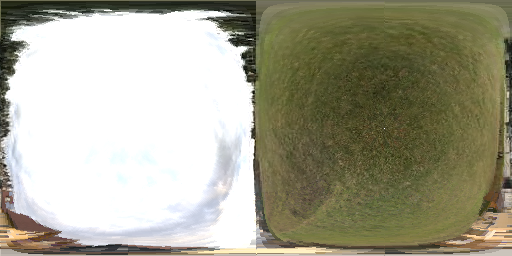}\\

{\rotatebox{90}{Ours}} &\includegraphics[width=0.19\textwidth]{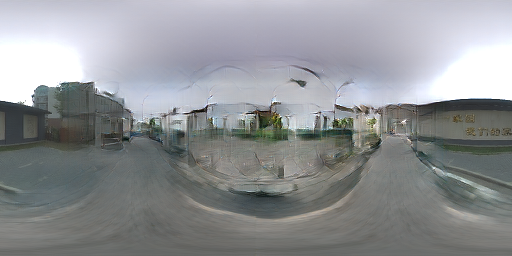}
&\includegraphics[width=0.19\textwidth]{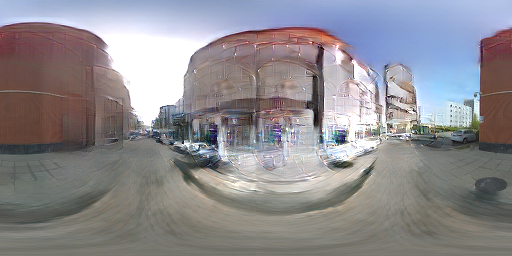} &\includegraphics[width=0.19\textwidth]{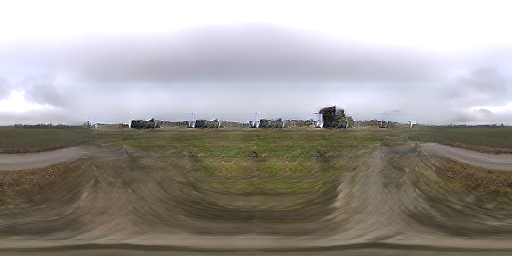} &\includegraphics[width=0.19\textwidth]{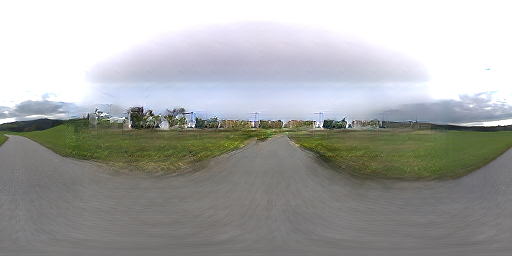} &\includegraphics[width=0.19\textwidth]{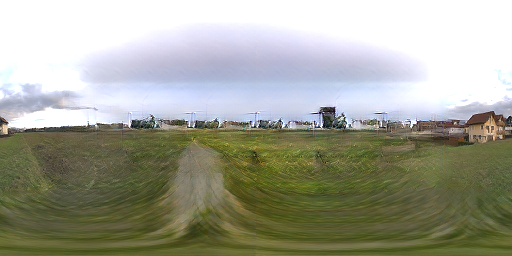}\\

&\includegraphics[width=0.19\textwidth]{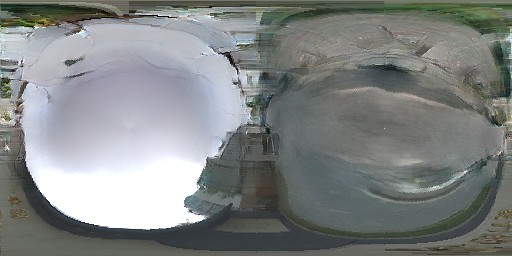}
&\includegraphics[width=0.19\textwidth]{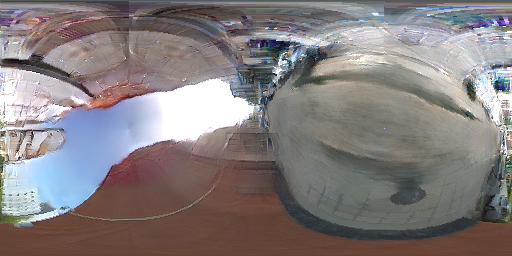} &\includegraphics[width=0.19\textwidth]{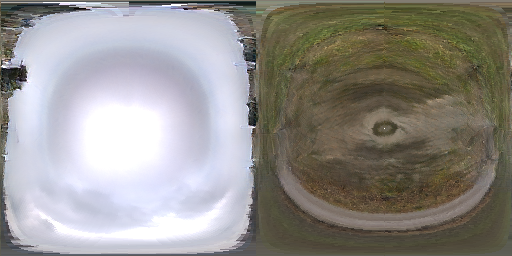} &\includegraphics[width=0.19\textwidth]{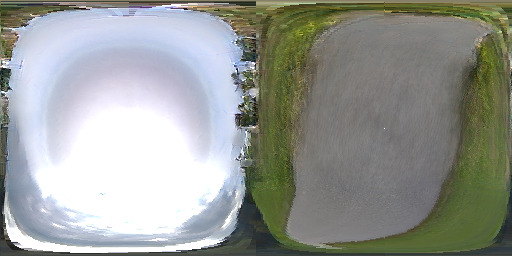} &\includegraphics[width=0.19\textwidth]{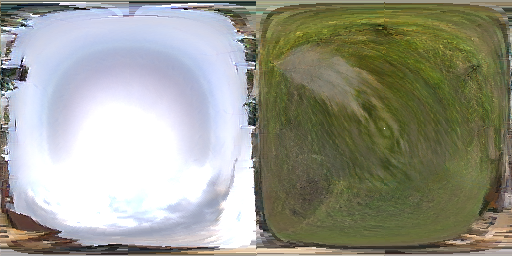}
\end{tabularx}
\end{center}
\caption{Outdoor qualitative comparison of our generated ERPs in LDR with other methods. For each method and input LFOV image we show the LDR ERP rotated 180$^\circ$ to show any potential border seams and the LDR ERP rotated 90$^\circ$ by 90$^\circ$ to compare the generation at the poles of the ERP.}
\label{fig:renders}
\end{figure}

In \cref{fig:indoor_renders}, we present a selection of panoramas rotated by 180$^\circ$ about the vertical axis to observe the side borders of the generated ERP. We also feature a rotation of 90$^\circ$ by 90$^\circ$ to show the poles of the panorama side by side. To be concise, we do not include all of the methods from the quantitative comparison and base our choice on the type of network and the quality of the output to prevent comparing against too many similar methods. These results show that our model completely removes the side border and any warping at the ERP's poles. Other methods are not able to remove the side border and, although it is not obvious, EverLight does not completely remove it. However, the plausibility of our generated textures and structures does not compete with the current state of the art. This reflects the quantitative results in \cref{T_Indoor}. Our method retains the information from the input LFOV LDRI as does EverLight.

\subsection{Outdoor Images}
\label{ssec:outdoor}

\begin{figure}[t]
\begin{center}
\setlength\tabcolsep{0pt}
\renewcommand{\arraystretch}{0.25}
\begin{tabularx}{\linewidth}{lXXXXX}

\rotatebox{90}{Input} &\centering \includegraphics[width=0.14\textwidth]{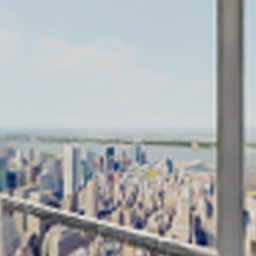} &\centering \includegraphics[width=0.14\textwidth]{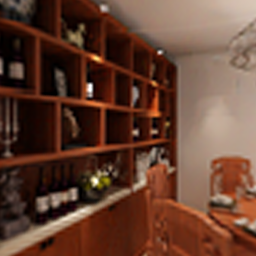} &\centering \includegraphics[width=0.14\textwidth]{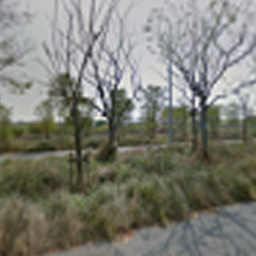} &\centering \includegraphics[width=0.14\textwidth]{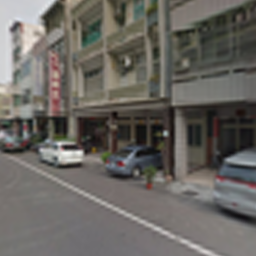} &\multicolumn{1}{c}{\includegraphics[width=0.14\textwidth]{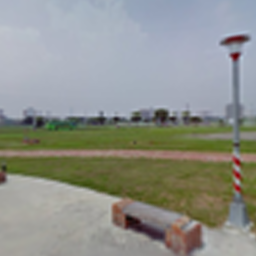}}\\

\multirow{2}*{\rotatebox{90}{Ground Truth  }}&&&&\\

&\includegraphics[width=0.19\textwidth]{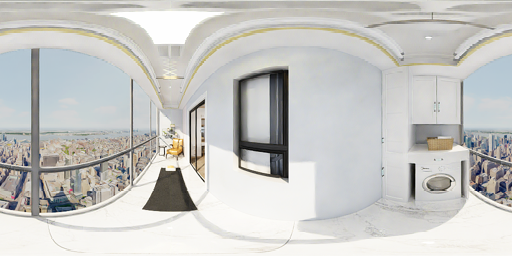}
&\includegraphics[width=0.19\textwidth]{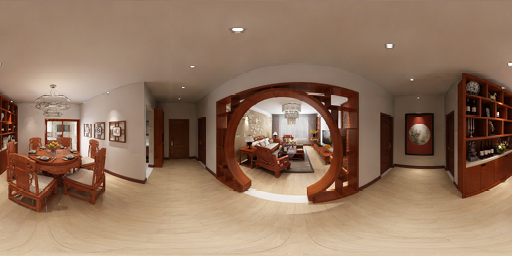}
&\includegraphics[width=0.19\textwidth]{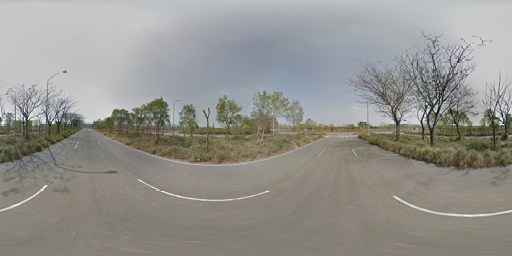} &\includegraphics[width=0.19\textwidth]{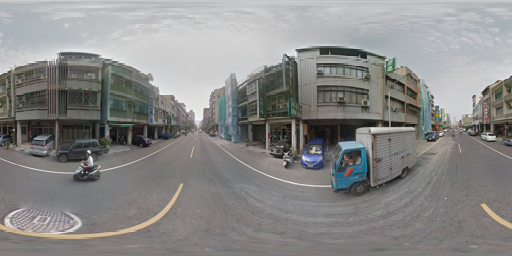} &\includegraphics[width=0.19\textwidth]{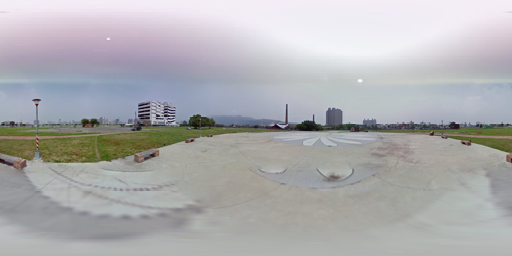}\\

\multirow{2}*{\rotatebox{90}{360U-Former  }} 
&\includegraphics[width=0.19\textwidth]{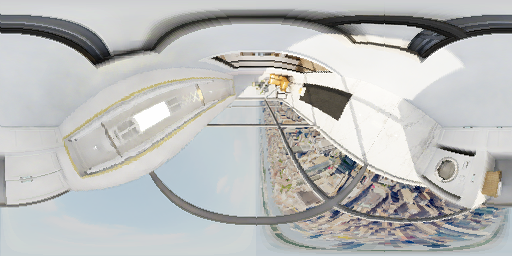}
&\includegraphics[width=0.19\textwidth]{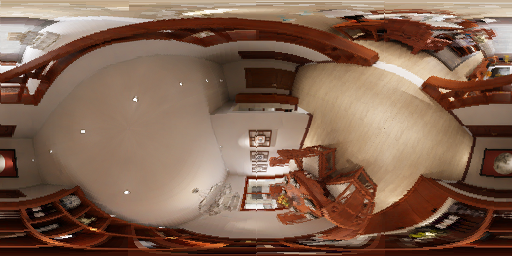}
&\includegraphics[width=0.19\textwidth]{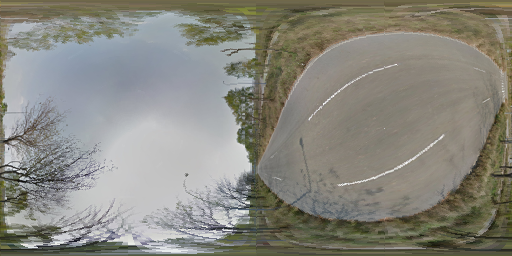} &\includegraphics[width=0.19\textwidth]{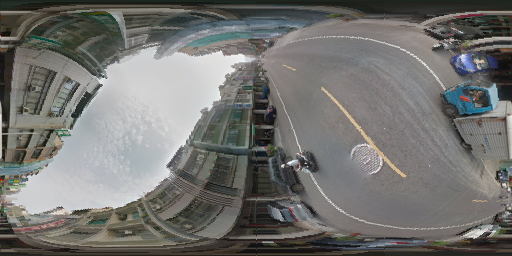} &\includegraphics[width=0.19\textwidth]{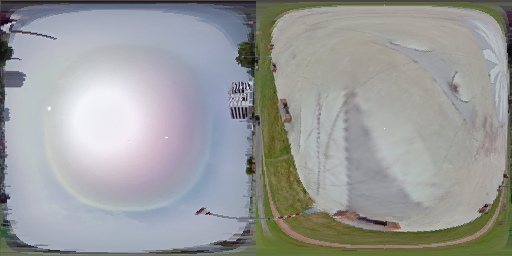}\\

&\includegraphics[width=0.19\textwidth]{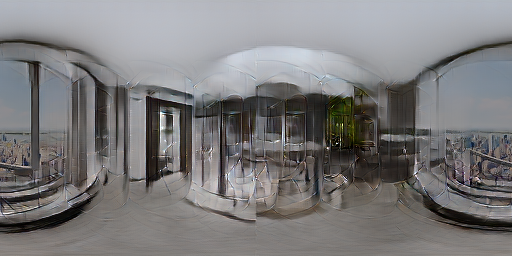}
&\includegraphics[width=0.19\textwidth]{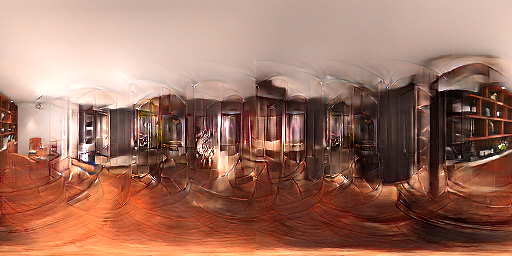}
&\includegraphics[width=0.19\textwidth]{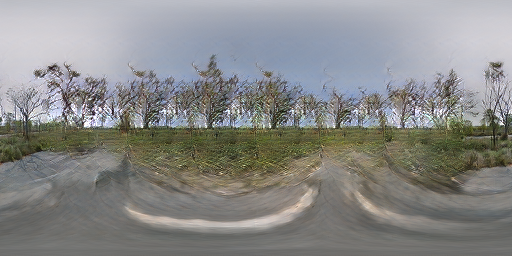} &\includegraphics[width=0.19\textwidth]{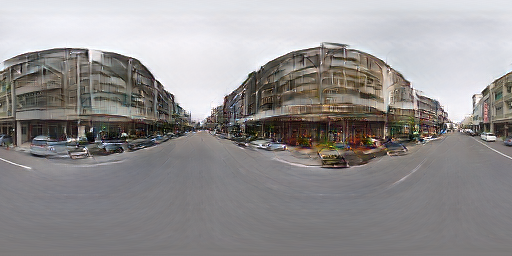} &\includegraphics[width=0.19\textwidth]{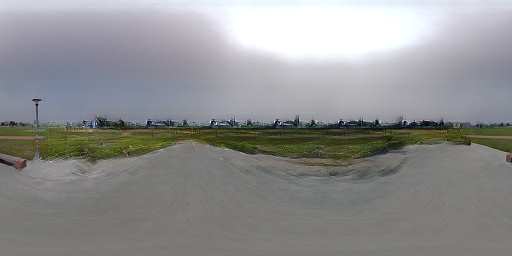}\\ 

\multirow{2}*{\rotatebox{90}{w/o PanoSWIN  }} 
&\includegraphics[width=0.19\textwidth]{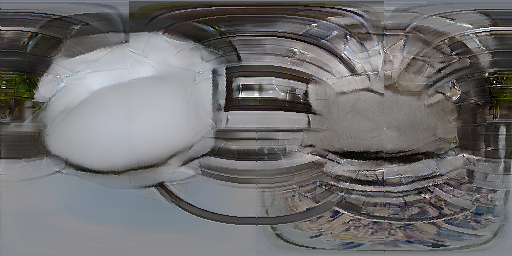}
&\includegraphics[width=0.19\textwidth]{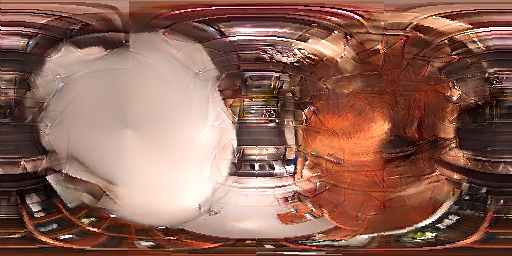}
&\includegraphics[width=0.19\textwidth]{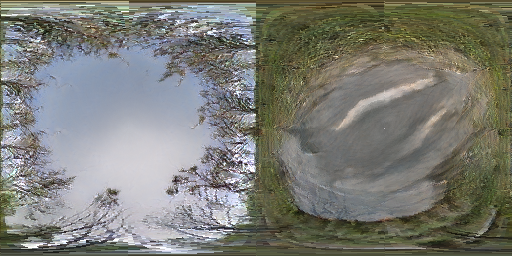} &\includegraphics[width=0.19\textwidth]{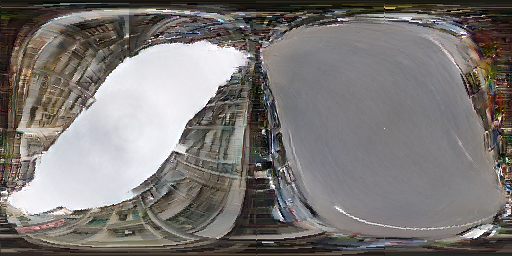} &\includegraphics[width=0.19\textwidth]{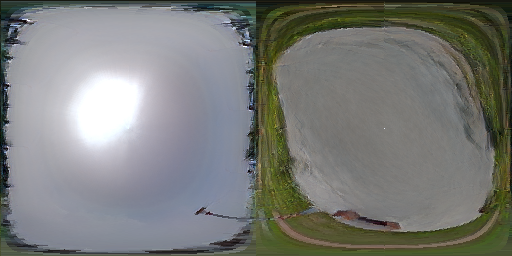}\\ 



&\includegraphics[width=0.19\textwidth]{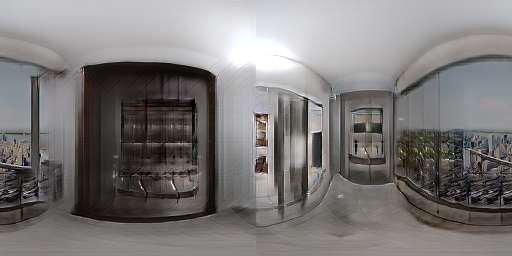}
&\includegraphics[width=0.19\textwidth]{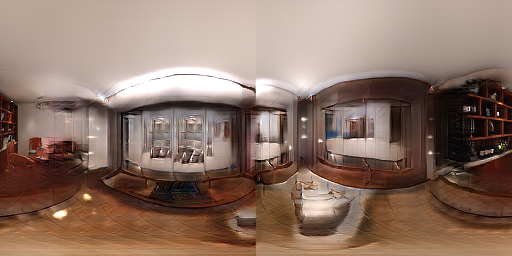}
&\includegraphics[width=0.19\textwidth]{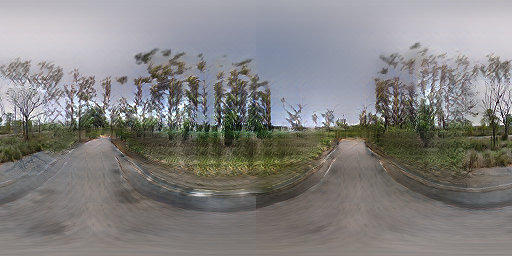} &\includegraphics[width=0.19\textwidth]{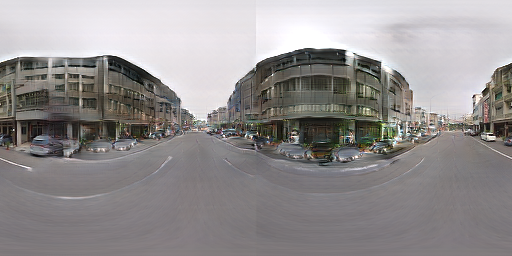} &\includegraphics[width=0.19\textwidth]{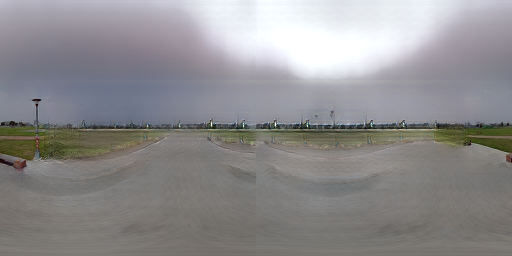}\\ 

&\includegraphics[width=0.19\textwidth]{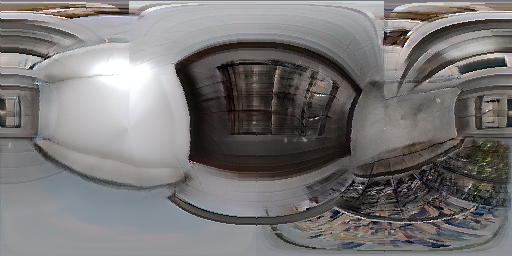}
&\includegraphics[width=0.19\textwidth]{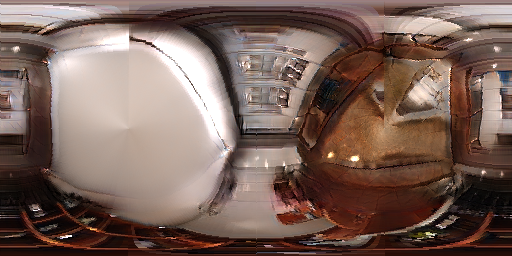}
&\includegraphics[width=0.19\textwidth]{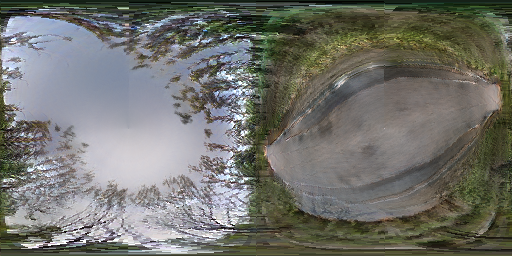} &\includegraphics[width=0.19\textwidth]{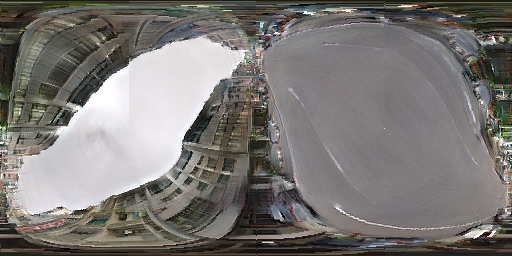} &\includegraphics[width=0.19\textwidth]{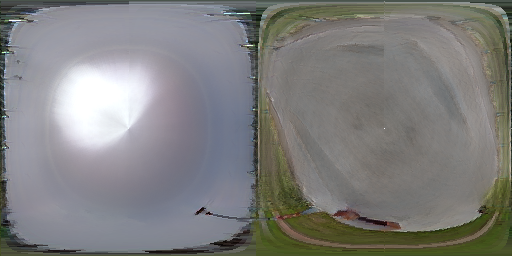}\\ 

\end{tabularx}
\end{center}
\caption{Qualitative comparison of our generated ERPs in LDR from our network with and without the PanoSWIN attention blocks. For each method and input LFOV image we show the LDR ERP rotated 180$^\circ$ to show any potential border seams and the LDR ERP rotated 90$^\circ$ by 90$^\circ$ to compare the generation at the poles of the ERP.}
\label{fig:ablation_renders}
\end{figure}

\subsubsection{Quantitative Results}
\label{sssec:outdoor_quant}

We conduct the quantitative comparison on the outdoor scenes similarly to the indoor scenes but change the input to 3 perspective crops at azimuth spacing \{0, 120 and 240\} of 90$^\circ$ field of view. We use the 893 outdoor panoramas from the \cite{Cheng2018} dataset giving a total of 2,517 images for evaluation. Unlike the indoor evaluation, all metrics are tested on this dataset. We compare our results with the works of Zhang \etal \cite{Zhang2019a}, Everlight \cite{Dastjerdi2023} and ImmerseGAN \cite{Dastjerdi2022}. It should be noted that Zhang \etal proposed a method for predicting the parameters of an outdoor sun+sky light model.

The quantitative results are shown in \cref{T_Indoor}. Similarly to the indoor comparison, our method outperforms all over methods when comparing the diffuse render showing a better understanding of light position and colour. As with the indoor method our FID scores perform worse than EverLight and ImmerseGAN, showing that our method does not reproduce plausible textures.

\subsubsection{Qualitative Results}
\label{sssec:outdoor_qual}

As with the indoor scenes we choose the same method to compare the quality of our generated images. We display images from the outdoor evaluation dataset from urban and natural scenes, similar to the ones in the 360 Sun Positions dataset. We also include some scene types from the evaluation dataset that are not included in our dataset to demonstrate our model’s ability to generalise. As with the indoor results our method removes the side border and warped poles commonly seen in illumination estimation methods. EverLight and ImmerseGAN have reduced the ERP warping to subtle artefacts only noticed when zooming in. We can see that the quality of the colours and features are similar to that of the ground truth. However, textures are not plausible and do not compete with the current state of the art. It is also worth noting that the model generates more plausible and accurate outdoor scenes with fewer artefacts compared to the indoor scenes.

\subsection{Ablation Study}
\label{ssec:ablate}

\begin{table}[t]
\caption{Ablation study to prove the effectivness of the PanoSWIN attention blocks in improving the models understanding of 360$^\circ$ ERP images. We highlight the best in {\bf bold}.}
\label{T_Ablation}
\centering
{\scriptsize
\begin{tabular}{|l|l|l|l|l|l|}
\hline
& Si-RMSE$\downarrow$ & RMSE$\downarrow$ & RGB ang.$\downarrow$ & PSNR$\uparrow$ & FID$\downarrow$ \\
\hline\hline
\multicolumn{6}{|l|}{INDOOR METHODS}\\
\hline
360U-Former &{\bf 0.033}&0.110&{\bf 6.11$^\circ$}&{\bf 11.68}&{\bf 119.91}\\
\hline
360U-Former with SWIN   &0.039&{\bf 0.104}&8.19$^\circ$&12.22&125.70\\
\hline
\multicolumn{6}{|l|}{OUTDOOR METHODS}\\
\hline
360U-Former &0.049&{\bf 0.161}&{\bf 4.00$^\circ$}&{\bf 13.27}&{\bf 102.63}\\

\hline
360U-Former with SWIN   &{\bf 0.047}&0.187&4.84$^\circ$&12.32&105.54\\
\hline
\end{tabular}
}
\end{table}

We conduct an ablation study to demonstrate the effectiveness of our adaptions, the PanoSWIN attention blocks and patch embedding with circular padding, at removing the artefacts produced when using the ERP format. We use the same methods of comparison as the indoor and outdoor quantitative and qualitative studies. The results can be seen in \cref{fig:ablation_renders} and \cref{T_Ablation} and show that not only does making adaptations to the network architecture remove the artefacts caused by using ERPs, it also improves the ability of a network to generate quality images. Most notable is the improvement in the RGB angular error, suggesting that adapting the network architecture has helped position light sources more accurately. The results shown in \cref{fig:ablation_renders} are from our test set. it should be noted that there is a significant improvement in the quality of the generation when our network uses images from the same dataset as the training set compared to the datasets used for the benchmark. This could be a limitation of the datasets we use to train the model, suggesting that there is not a diverse enough distribution of data in the Structured3D \cite{Zheng2019} and 360 Sun Positions \cite{Chang2018} datasets.

\section{Conclusion}
\label{sec:Conclusion}

This paper proposes a method for removing artefacts caused by using the ERP format and a Vision-Transformer architecture for estimating the illumination conditions of indoor and outdoor scenes from an LFOV image. By utilising a Vision-Transformer architecture with PanoSWIN attention layers the network can account for the warping at the poles of the ERP and allow for seamless and homogenised generation at the borders. We demonstrate this through a qualitative comparison that rotates the generated environment maps, highlighting warping at the edges of the image. In general, the results of this paper could be used as a guide when constructing any neural network that makes use of the ERP image format to improve the quality of the generation and remove artefacts.
Although the ERP artefacts have been addressed, the network lacks the ability to generate plausible high-resolution textures competitive with that of other state-of-the-art methods. This could potentially be resolved by increasing the size of the dataset or using a different training method.

Further analysis could be carried out by mapping the attention and latent space to understand how the PanoSWIN attention layers improve the ViT network. We could also compare our models' complexity with other approaches.
Based on the direction the field of illumination estimation is heading towards we highlight three key areas that could extend the functionality of our method. Firstly it would be useful for a user to have a variety of mediums to edit the generated output, such as providing a diffuse environment map to the ICN to change the lighting conditions or by using a text prompt to change the details of the environment map. Secondly, adapting the method to incorporate spatial variance to shift the panorama so that it can accurately light objects not at its centre. This feature would be integral to XR applications. Finally, a large-scale standardised HDR panoramic dataset that has calibrated values for light sources and a range of outdoor and indoor scenes with various textures.

\section*{Acknowledgements}
\label{sec:Acknowledgements}

This work was supported by the UKRI EPSRC Doctoral Training Partnership Grants EP/N509772/1 and EP/R513350/1 (studentship reference 2437074)

\clearpage

%
%
\bibliographystyle{splncs04}

\end{document}